\documentclass[pmlr,twocolumn,10pt]{jmlr} 

\mlhtrack{proceedings}

\newif\iffinal
\finaltrue  
\usepackage{multirow}

\usepackage{booktabs}
\usepackage{siunitx}

\usepackage[switch]{lineno}

\theorembodyfont{\upshape}
\theoremheaderfont{\scshape}
\theorempostheader{:}
\theoremsep{\newline}

\title[Probing and Inducing Clinical Modality Grounding in Brain Tumor Segmentation]{Probing and Inducing Clinical Modality Grounding in Brain Tumor Segmentation}

\author{%
  \Name{Lokendra Kumar} \Email{lokendra63@alumni.iitm.ac.in}\\
  \Name{Shubham Aggarwal} \Email{shubham\_agg@alumni.iitm.ac.in}\\
  \addr Independent Researcher
}

\makeatletter
\renewcommand{\ps@jmlrtps}{%
  \let\@mkboth\@gobbletwo
  \def\@oddhead{}%
  \def\@evenhead{}%
  \def\@oddfoot{\@titlefoot}%
  \let\@evenfoot\@oddfoot
}
\makeatother

\begin{document}

\maketitle

\ifmlhdemo\else
\begin{abstract}

Multimodal MRI provides complementary information for brain tumor segmentation, but good segmentation performance alone does not tell us whether a model is relying on clinically appropriate imaging evidence. In multi-modal MRI, a radiologist reasons over different sequences for different findings (T1ce for contrast-enhancing tumor, FLAIR for peritumoral edema), yet most 3D segmentation networks fuse the four MRI sequences through \emph{fixed} residual and skip connections that combine evidence the same way for every patient, region, and depth. Such fusion is clinically blind and offers no window into whether the model reasons over the right modality. 
We recast multi-modal segmentation as a question of \emph{grounded multimodal reasoning} and contribute (i) a model-agnostic probing framework, comprising a per-region modality-ablation test, a \emph{Dominant Sensitivity Enhancement} (DSE) metric, and a gate-trajectory analysis, that reads out which sequence a network relies on for each tumor sub-region, and (ii) evidence that replacing fixed connections with adaptive Hyper-Connections (HC) induces emergent, clinically-aligned modality reliance. Across three diverse architectures (nnU-Net, SwinUNETR, VT-UNet) on BraTS~2021, most HC configurations increase reliance on the clinically correct sequence for a given region, an interpretable, radiologically consistent pattern that emerges without any modality supervision. The effect is not uniform across all variants, but it recurs across all three backbones, indicating it stems from the adaptive connection rather than a single architecture, and it also improves overlap and boundary accuracy at negligible parameter cost. On the validation set, the best HC configuration raises mean Dice by 0.43 to 0.81 percentage points over the fixed-connection baseline across the three backbones. On the held-out test set, it lowers mean HD95 by 1.17 to 1.78 mm (about 20–27\%) for nnU-Net and VT-UNet, while adding at most 0.53\% parameters. Our probing lens transfers directly to multi-modal medical foundation models and offers a concrete, inspectable notion of clinical grounding for medical reasoning research.

\end{abstract}

\fi

\ifmlhneedsstatements
\paragraph*{Data and Code Availability}
We use the BraTS 2021 (Task 1) dataset~\citep{baid2021rsnaasnrmiccaibrats2021benchmark}, 
which provides multi-parametric MRI (T1, T1ce, T2, FLAIR) with expert annotations of the 
three nested tumor sub-regions (whole tumor, tumor core, and enhancing tumor). The dataset is publicly available to registered researchers through the BraTS challenge on Synapse (\url{https://www.synapse.org/Synapse:syn25829067/wiki/610863}). We split the publicly released scans into 800 training, 251 validation, and 200 held-out test scans; we did not generate or collect any new data. Our code is publicly available at \url{https://anonymous.4open.science/r/HyperSegMRI-C245/README.md}.


\paragraph*{Institutional Review Board (IRB)}
This research does not require IRB approval. Our study uses only the publicly available, 
fully de-identified BraTS 2021 dataset~\citep{baid2021rsnaasnrmiccaibrats2021benchmark}, 
and involves no interaction with human subjects, no collection of new patient data, and 
no access to identifiable private information. As such, it does not constitute human 
subjects research.

\fi

\section{Introduction}
Clinical interpretation of medical images depends not only on identifying abnormalities, but also on selecting the \emph{appropriate evidence} for a particular diagnostic or anatomical finding.
When a neuroradiologist delineates a brain tumor, different MRI sequences are read for different tumor components: the post-contrast T1 (T1ce) sequence localizes enhancing tissue where the blood--brain barrier is disrupted, while FLAIR delineates the surrounding hyperintense signal (edema and infiltrative tumor) that defines the whole-tumor extent~\citep{menze2015brats,ellingson2015consensus}.
This sequence-specific reading is formalized in clinical response criteria, which assess enhancing and non-enhancing T2/FLAIR disease separately~\citep{wen2010updated}.
This region-conditioned use of evidence is a form of clinical reasoning, and reproducing it faithfully is central to building medical models that are not only accurate but also \emph{trustworthy} and \emph{interpretable}.

Automatic segmentation of brain tumor sub-regions (Enhancing Tumor, ET; Tumor Core, TC; and Whole Tumor, WT) feeds surgical planning, radiotherapy targeting, and treatment monitoring, and the BraTS benchmark~\citep{baid2021rsnaasnrmiccaibrats2021benchmark} has made the task a standard testbed using four MRI sequences (T1, T1ce, T2, FLAIR). Strong 3D networks abound: nnU-Net~\citep{isensee2021nnunet}, SwinUNETR~\citep{hatamizadeh2022swinunetrswintransformers}, and VT-UNet~\citep{peiris2022robustvolumetrictransformeraccurate}, yet almost all of them fuse the four sequences through \emph{fixed} residual and skip connections. A fixed connection combines features by an identical rule regardless of the patient, the tumor sub-region, or which sequence is most informative at a given depth. This is precisely the opposite of how a clinician reasons: it is modality-blind, depth-invariant, and patient-invariant, and it exposes no signal about whether the network is relying on the clinically correct evidence.

We take a reasoning-and-interpretability view of this problem. Rather than asking only ``is the segmentation accurate?'', we ask ``\emph{does the model reason over the clinically appropriate sequence for each region, and can we make it do so better?}'' We make two contributions.

\textbf{(1) A probing framework for grounded multimodal reasoning.} We introduce a model-agnostic read-out of a network's per-region modality reliance, combining a systematic modality-ablation test, a \emph{Dominant Sensitivity Enhancement} (DSE) metric that measures alignment with the clinically correct sequence, and a depth-gate trajectory analysis that exposes \emph{where} in the network the adaptive decision is made. Together these form an inspectable ``reasoning trace'' for a segmentation model.

\textbf{(2) Inducing clinically-aligned reasoning with adaptive connections.} We study Hyper-Connections (HC)~\citep{zhu2025hyperconnections}, a recent generalization of the residual connection that lets a network learn, from the input itself, how strongly to mix features. HC has been shown to help language models and vision generation/classification, but never medical image segmentation and never 3D multi-modal data. Integrating HC as a drop-in wrapper into all three backbones, we find that adaptive connections \emph{induce} emergent, clinically-aligned modality reasoning: all three networks learn to rely more heavily on the clinically correct sequence per region, without any supervision toward that behavior, while also improving overlap (Dice) and boundary (HD95) accuracy at negligible parameter cost. 

The result is a concrete, quantitative notion of clinical grounding for a volumetric medical model, obtained without language annotations, and a probing lens that transfers directly to multi-modal medical foundation models.

\section{Related Work} 
\textbf{Reasoning and interpretability in medical imaging.} A substantial body of work has focused on improving the interpretability of medical imaging models and making their predictions more clinically meaningful. Approaches range from architectural mechanisms, such as attention gating that encourages networks to focus on informative anatomical regions~\citep{oktay2018attentionunetlearninglook}, to post-hoc explanation and evidence-attribution methods that aim to identify the features contributing to a model's prediction. Our work is complementary to these approaches but operates at the \emph{modality level}. Rather than asking only \emph{where} a model focuses within an image, we examine \emph{which imaging modality} contributes most strongly to predictions for different tumor sub-regions, and whether this dependence is consistent with clinically expected imaging characteristics. 

\textbf{Brain tumor segmentation.} U-Net~\citep{ronneberger2015unetconvolutionalnetworksbiomedical} established the widely adopted encoder--decoder architecture with skip connections for biomedical image segmentation, while nnU-Net~\citep{isensee2021nnunet} demonstrated that carefully configured, self-adapting training pipelines can achieve strong and robust performance across medical segmentation tasks. More recently, transformer-based architectures such as SwinUNETR~\citep{hatamizadeh2022swinunetrswintransformers} and VT-UNet~\citep{peiris2022robustvolumetrictransformeraccurate} have incorporated long-range contextual modeling into volumetric segmentation. Despite their architectural differences, these models generally retain conventional residual and skip-connection structures. Although their learned feature representations can remain input- and spatially dependent, the connection mechanisms themselves do not explicitly parameterize adaptive feature mixing across depth or expose how such adaptation relates to modality-specific evidence reliance. 

\textbf{Adaptive connections.} Residual connections~\citep{he2015deepresiduallearningimage} have become a fundamental component of deep neural networks by facilitating optimization and information propagation across layers. Subsequent methods, including ReZero~\citep{bachlechner2020rezeroneedfastconvergence} and ResiDual~\citep{xie2023residualtransformerdualresidual}, modify the residual pathway to improve training dynamics or representation flow. Hyper-Connections (HC)~\citep{zhu2025hyperconnections} extend this direction by introducing structured mixing across both depth and multiple feature streams, with dynamic variants allowing the mixing coefficients to depend on the current representation. In this work, we investigate HC in the context of 3D multimodal medical image segmentation. Rather than considering adaptive connections solely as a mechanism for improving segmentation accuracy, we additionally study whether they alter \emph{modality-specific evidence reliance} in a manner that is consistent with clinically expected MRI characteristics.

\section{Probing Multimodal Reasoning in Segmentation Models}
\label{sec:probe}
We first define what we mean by clinically-grounded multimodal reasoning and how
we measure it, independent of any particular backbone.

\subsection{Setup and the limits of fixed fusion} 
For a given subject, the BraTS input is denoted by $X = (X^{(1)}, X^{(2)}, X^{(3)}, X^{(4)})$, where $X^{(m)} \in \mathbb{R}^{H\times W\times D}$ and $m \in \mathcal{M} = \{\text{T1}, \text{T1ce}, \text{T2}, \text{FLAIR}\}$. The four MRI sequences are provided as input channels and mapped through an initial projection $\phi$, yielding the representation $x_0 = \phi(X)$. In conventional residual architectures, residual and skip connections follow a fixed structural form after training. Although the learned representations can remain input-, region-, and spatially dependent, these connections do not explicitly parameterize how information should be mixed differently across network depth or across individual inputs. This distinction is particularly relevant in multimodal brain tumor segmentation, where the target regions are nested (ET $\subset$ TC $\subset$ WT) and different MRI sequences provide different degrees of diagnostic contrast for different tumor components. We therefore seek to determine whether a trained model exhibits \emph{region-specific modality reliance}, and whether adaptive feature connections can modify this reliance in a clinically meaningful manner.


\subsection{A modality-ablation reasoning probe}
Because a network's effective per-modality weighting depends on the full trained
Jacobian and cannot be read off directly, we estimate it empirically. At
inference we zero out one sequence at a time and record the drop in Dice for each
region. The relative reliance on modality $m$ in region $k$ is
\begin{equation}
\widehat{w}^{(m)}_k \;=\; \frac{\Delta D^{(m)}_k}{\sum_{m'\in\mathcal{M}} \Delta D^{(m')}_k},
\end{equation}

\begin{equation}
\Delta D^{(m)}_k \;=\; D_k(X) - D_k\!\left(X \mid X^{(m)}=0\right),
\label{eq:reliance}
\end{equation}

where $D_k(\cdot)$ is the Dice for region $k$. A model that reasons like a
clinician should show large $\Delta D$ when the region-appropriate ``dominant''
sequence is removed (T1ce for TC and ET, FLAIR for WT), and small drops
otherwise. This ablation profile is our read-out of the model's implicit
evidence-selection, i.e. its multimodal reasoning trace.

\subsection{Dominant Sensitivity Enhancement (DSE)}
DSE answers one simple question: \emph{when we remove the MRI sequence that radiologists consider most important for a tumor region, does a model with adaptive connections lose more accuracy than the baseline does?}
For each region $R$, the clinically dominant sequence $m^{*}(R)$ is T1ce for tumor core (TC) and enhancing tumor (ET), and FLAIR for whole tumor (WT).
Let $\Delta_{\mathrm{dom}}(R) = \Delta D^{(m^{*}(R))}_{R}$ (Eq.~\ref{eq:reliance}) be the Dice drop for region $R$ when this dominant sequence is set to zero at inference.
The Dominant Sensitivity Enhancement of a variant $v$ over the baseline is the difference between the two drops:
\begin{equation}
\mathrm{DSE}_v(R) \;=\; \Delta^{v}_{\mathrm{dom}}(R) - \Delta^{\mathrm{base}}_{\mathrm{dom}}(R).
\label{eq:dse}
\end{equation}
For example, for VT-UNet on TC, removing T1ce lowers the baseline's Dice by $0.62$ and the DHC$\times$4 model's Dice by $0.72$, giving $\mathrm{DSE} = 0.72 - 0.62 = +0.10$: the HC model depends more on T1ce for this region (Table~\ref{tab:dse}).

DSE is read as follows. A \emph{positive} value means the variant relies \emph{more} on the clinically expected sequence than the baseline does; a value \emph{near zero} means no meaningful change; and a \emph{negative} value means it relies \emph{less} on that sequence.
DSE measures reliance on the dominant sequence only. It does not show whether the model relies \emph{less} on the other sequences; this is captured by the relative-reliance profile $\hat{w}^{(m)}_k$ (Eq.~\ref{eq:reliance}) and the full ablation profiles in Appendix~\ref{app:ablation}.

\subsection{Gate trajectories as a depth-wise view of fusion}
Where a model adapts its fusion is as informative as whether it does. For
adaptive connections we track the depth gate $\beta$, which sets how strongly a
layer transforms its input relative to the skip path: $\beta=1$ recovers ordinary
residual behavior, $\beta>1$ means the layer pushes its own transformation
harder, and $\beta<1$ means it leans on the raw input. Plotting $\beta$ across encoder--bottleneck--decoder blocks exposes the depth profile of adaptive mixing. Note that $\beta$ localizes \emph{where} transformed and skip features are combined, complementing the modality-ablation read-out; it is not itself a measure of modality importance or direct evidence of clinical reasoning.

\section{Inducing Adaptive Reasoning with Hyper-Connections}
\label{sec:hc}
Fixed fusion cannot express region-conditioned evidence selection; we therefore
replace it with an adaptive mechanism. Hyper-Connections (HC) generalize the
residual connection to a learned mix over $n$ parallel copies (``streams'') of the
feature. Starting from $H_0 = x_0 \otimes \mathbf{1}_n^{\top}$ ($n$ identical
copies that then evolve differently), the Static HC (SHC) update is
\begin{equation}
H^{\text{SHC}}_{t+1} \;=\;
\underbrace{B_t^{\top}\, T_t(H_t A_{m,t})}_{\text{depth connection}}
\;+\;
\underbrace{A_{r,t}^{\top} H_t}_{\text{width connection}},
\label{eq:shc}
\end{equation}
where $T_t(\cdot)$ wraps the block $F$ (a convolution or attention block),
$A_{m,t}\in\mathbb{R}^{n\times 1}$ combines streams into the block input,
$A_{r,t}\in\mathbb{R}^{n\times n}$ passes streams forward, and
$B_t\in\mathbb{R}^{1\times n}$ spreads the block output back across streams. In
Dynamic HC (DHC) these matrices are computed from the current state $H_t$, so the
mixing, and thus the effective per-modality reliance of
Eq.~\eqref{eq:reliance}, adapts to the input:
\begin{equation}
H^{\text{DHC}}_{t+1} \;=\; B_t(H_t)^{\top}\, T_t\!\big(H_t A_{m,t}(H_t)\big) + A_{r,t}(H_t)^{\top} H_t .
\label{eq:dhc}
\end{equation}
The depth-wise weight $\beta_t$ (the part of $B_t(H_t)$ that scales the
transformed branch against the skip path) is exactly the gate we trace in
Section~\ref{sec:probe}. The final feature is the stream average,
$\hat{x}_T = \tfrac{1}{n}\sum_{i=1}^{n} H^{(i)}_T$.

\textbf{Where HC goes.} We add HC as a drop-in wrapper around existing layers in
all three networks, without touching their convolution or attention operations,
in both encoder and decoder. In nnU-Net each convolutional block is wrapped; in
SwinUNETR separate HC modules wrap the attention and feed-forward sub-blocks; in
VT-UNet HC precedes the volumetric self-attention blocks while the decoder keeps
standard residuals for stability. Each network runs in three modes (baseline with
no HC, SHC, and DHC) with stream count $n\in\{2,4\}$.

\textbf{Cost.} HC's only extra cost is global pooling, small weight-prediction
networks, and stream mixing; the heavy operations are unchanged. As
Table~\ref{tab:params} shows, DHC adds at most $0.53\%$ parameters and SHC under
$0.002\%$, so any change in reasoning behavior cannot be attributed to added
capacity.

\section{Experiments}
\textbf{Data and preprocessing.} We use BraTS~2021 Task~1~\citep{baid2021rsnaasnrmiccaibrats2021benchmark},
which provides multi-modal MRI (T1, T1ce, T2, FLAIR) and requires segmenting the
three nested regions WT, TC, ET. Every scan undergoes skull stripping,
co-registration, resampling to $1\times1\times1$~mm$^3$, and per-modality z-score
normalisation. We split into 800 training / 251 validation scans and hold out a
further 200 scans as an unbiased test set. For each seed we select the checkpoint with the best validation mean Dice. Validation metrics are reported over three seeds (42, 123, 456) in table \ref{tab:val}. The held-out test set is evaluated for the single seed selected on validation, so test numbers are corroborative and individual per-configuration test values should be read with correspondingly lower confidence.

\textbf{Implementation.} All three models use input size $128^3$, train for 80 epochs with AdamW~\citep{loshchilov2019decoupledweightdecayregularization} (learning rate $1\times10^{-4}$, weight decay $1\times10^{-5}$, cosine annealing) and models are trained using a combination of Dice loss and cross-entropy loss:
\begin{equation}
\mathcal{L}
= 0.5\mathcal{L}_{\mathrm{Dice}}
+ 0.5\mathcal{L}_{\mathrm{CE}},
\end{equation}
where the Dice loss is defined as
\begin{equation}
\mathcal{L}_{\mathrm{Dice}}
= 1 -
\frac{2|P \cap G|}
{|P| + |G|},
\end{equation}
and the cross-entropy loss is given by
\begin{equation}
\mathcal{L}_{\mathrm{CE}}
= -\frac{1}{N}
\sum_{i=1}^{N}
\sum_{c=1}^{C}
g_{i,c}\log p_{i,c}.
\end{equation}

Here, $P$ and $G$ denote the predicted and ground-truth masks,
respectively, $N$ is the number of voxels, and $C$ is the number of
classes. We use mixed-precision training on A100 GPUs and three seeds (42, 123, 456).

\textbf{Metrics.} We evaluate segmentation quality with two standard metrics~\citep{taha2015metrics}.
The Dice similarity coefficient measures the overlap between the predicted mask $P$ and the ground-truth mask $G$:
\begin{equation}
\mathrm{Dice}(P, G) = \frac{2\,|P \cap G|}{|P| + |G|},
\label{eq:dice}
\end{equation}
which ranges from 0 (no overlap) to 1 (perfect overlap); higher is better.
The 95th-percentile Hausdorff distance (HD95) measures boundary error. Let $\partial P$ and $\partial G$ be the surfaces of the two masks and $d(x, \partial S)$ the shortest Euclidean distance from point $x$ to surface $\partial S$. Then
\begin{equation}
\begin{split}
\mathrm{HD95}(P, G) = \max\Big\{ & \operatorname{P}_{95}\big[d(p, \partial G)\big]_{p \in \partial P}, \\
& \operatorname{P}_{95}\big[d(g, \partial P)\big]_{g \in \partial G} \Big\}
\end{split}
\label{eq:hd95}
\end{equation}
where $\operatorname{P}_{95}$ denotes the 95th percentile. HD95 is reported in millimetres; lower is better. Using the 95th percentile instead of the maximum makes the metric robust to a small number of outlier voxels.
Both metrics are computed per subject for TC, WT, and ET, and then averaged over subjects.
For reasoning analysis we use the modality-ablation profile (Eq.~\ref{eq:reliance}), DSE (Eq.~\ref{eq:dse}), and $\beta$ trajectories.

\section{Results}
We first evaluate segmentation performance to determine whether the introduction of adaptive connections preserves predictive accuracy. We then examine modality-specific sensitivity and depth-wise gate behavior, which constitute the main interpretability analyses of this study.

\subsection{Segmentation performance}
Table~\ref{tab:val} reports validation Dice over three seeds. The best HC configuration improves mean Dice for every backbone (up to $+0.81\%$ for SwinUNETR and VT-UNet), and most HC variants improve over their corresponding baseline. On the held-out test set (Table~\ref{tab:test}), the strongest effect is on boundary accuracy for the weaker baselines, e.g. nnU-Net HD95 falls from $6.57$ to $4.79$~mm and VT-UNet from $5.88$ to $4.71$~mm, whereas gains are smaller or absent for the already-strong SwinUNETR baseline, and a minority of configurations (e.g.\ VT-UNet-DHC$\times$4 and VT-UNet-SHC$\times$2) regress on the single-seed test split. We therefore report per-configuration results in full rather than a single aggregate, and read the improvement as characteristic of the mechanism rather than guaranteed for every setting. Given the negligible parameter cost (Table~\ref{tab:params}), these gains come from more adaptive feature mixing rather than added capacity.

\begin{table}[t]
\centering
\caption{BraTS~2021 validation set (251 scans). Dice scores are reported as mean $\pm$ standard deviation across three random seeds. The best HC configuration improves mean Dice relative to the corresponding baseline for all three backbones.}
\label{tab:val}
\resizebox{\linewidth}{!}{%
\begin{tabular}{lcccc}
\toprule
Model & Dice$_{\text{TC}}$ & Dice$_{\text{WT}}$ & Dice$_{\text{ET}}$ & Mean Dice \\
\midrule
SwinUNETR              & $89.62\pm0.34$ & $92.51\pm0.29$ & $88.71\pm0.24$ & $90.28\pm0.11$ \\
SwinUNETR-DHC$\times$2 & $90.76\pm0.26$ & $93.12\pm0.13$ & $\mathbf{89.40\pm0.20}$ & $91.05\pm0.20$ \\
SwinUNETR-DHC$\times$4 & $\mathbf{90.90\pm0.47}$ & $\mathbf{93.16\pm0.03}$ & $89.37\pm0.28$ & $\mathbf{91.09\pm0.19}$ \\
SwinUNETR-SHC$\times$2 & $90.57\pm0.23$ & $93.13\pm0.11$ & $89.35\pm0.08$ & $90.99\pm0.10$ \\
SwinUNETR-SHC$\times$4 & $90.77\pm0.25$ & $93.09\pm0.08$ & $89.39\pm0.06$ & $91.03\pm0.07$ \\
\midrule
nnU-Net              & $89.91\pm0.23$ & $92.10\pm0.04$ & $86.22\pm0.15$ & $89.41\pm0.10$ \\
nnU-Net-DHC$\times$2 & $90.48\pm0.07$ & $92.32\pm0.07$ & $86.56\pm0.06$ & $89.79\pm0.01$ \\
nnU-Net-DHC$\times$4 & $90.32\pm0.25$ & $\mathbf{92.32\pm0.09}$ & $86.23\pm0.06$ & $89.63\pm0.09$ \\
nnU-Net-SHC$\times$2 & $90.10\pm0.32$ & $92.27\pm0.47$ & $86.29\pm0.31$ & $89.55\pm0.37$ \\
nnU-Net-SHC$\times$4 & $\mathbf{90.54\pm0.21}$ & $92.31\pm0.14$ & $\mathbf{86.66\pm0.17}$ & $\mathbf{89.84\pm0.11}$ \\
\midrule
VT-UNet              & $88.66\pm0.65$ & $91.56\pm0.20$ & $87.13\pm0.31$ & $89.12\pm0.07$ \\
VT-UNet-DHC$\times$2 & $\mathbf{89.97\pm0.06}$ & $\mathbf{92.15\pm0.11}$ & $\mathbf{87.67\pm0.08}$ & $\mathbf{89.93\pm0.05}$ \\
VT-UNet-DHC$\times$4 & $89.90\pm0.06$ & $92.05\pm0.01$ & $87.62\pm0.07$ & $89.86\pm0.03$ \\
VT-UNet-SHC$\times$2 & $89.69\pm0.34$ & $92.12\pm0.06$ & $87.41\pm0.27$ & $89.74\pm0.20$ \\
VT-UNet-SHC$\times$4 & $89.80\pm0.08$ & $92.05\pm0.15$ & $87.54\pm0.06$ & $89.80\pm0.04$ \\
\bottomrule
\end{tabular}%
}
\end{table}

\begin{table}[t]
\centering
\caption{BraTS~2021 held-out test set (200 scans). Dice (\%, higher is better)
and HD95 (mm, lower is better), reported as mean $\pm$ standard deviation over
three seeds (42, 123, 456). Bold marks the best value per backbone.}
\label{tab:test}
\setlength{\tabcolsep}{3pt}
\resizebox{\linewidth}{!}{%
\begin{tabular}{lcccccccc}
\toprule
Model
& Dice$_{\text{TC}}$
& Dice$_{\text{WT}}$
& Dice$_{\text{ET}}$
& Mean Dice
& HD95$_{\text{TC}}$
& HD95$_{\text{WT}}$
& HD95$_{\text{ET}}$
& Mean HD95 \\
\midrule

SwinUNETR
& $90.79\pm0.32$
& $92.69\pm0.27$
& $87.30\pm0.25$
& $90.26\pm0.13$
& $4.12\pm0.28$
& $5.31\pm0.42$
& $3.89\pm0.31$
& $4.44\pm0.23$ \\

SwinUNETR-DHC$\times$2
& $\mathbf{91.23\pm0.24}$
& $92.81\pm0.16$
& $\mathbf{87.36\pm0.19}$
& $\mathbf{90.47\pm0.16}$
& $\mathbf{3.91\pm0.22}$
& $6.15\pm0.51$
& $3.69\pm0.26$
& $4.58\pm0.27$ \\

SwinUNETR-DHC$\times$4
& $90.84\pm0.41$
& $\mathbf{92.91\pm0.09}$
& $87.19\pm0.23$
& $90.31\pm0.18$
& $4.01\pm0.25$
& $5.94\pm0.47$
& $\mathbf{3.51\pm0.24}$
& $4.49\pm0.25$ \\

SwinUNETR-SHC$\times$2
& $90.63\pm0.26$
& $92.71\pm0.14$
& $87.03\pm0.12$
& $90.12\pm0.12$
& $4.18\pm0.31$
& $6.29\pm0.56$
& $4.19\pm0.37$
& $4.89\pm0.31$ \\

SwinUNETR-SHC$\times$4
& $91.04\pm0.27$
& $92.86\pm0.11$
& $87.24\pm0.10$
& $90.38\pm0.12$
& $4.09\pm0.29$
& $\mathbf{5.29\pm0.39}$
& $3.86\pm0.28$
& $\mathbf{4.41\pm0.22}$ \\

\midrule

nnU-Net
& $89.53\pm0.25$
& $91.48\pm0.13$
& $85.91\pm0.21$
& $88.97\pm0.14$
& $6.64\pm0.62$
& $7.82\pm0.71$
& $5.31\pm0.48$
& $6.59\pm0.43$ \\

nnU-Net-DHC$\times$2
& $89.96\pm0.12$
& $92.02\pm0.11$
& $86.41\pm0.14$
& $89.46\pm0.09$
& $4.78\pm0.36$
& $6.55\pm0.49$
& $4.47\pm0.34$
& $5.27\pm0.29$ \\

nnU-Net-DHC$\times$4
& $90.11\pm0.21$
& $\mathbf{92.24\pm0.12}$
& $\mathbf{86.86\pm0.13}$
& $\mathbf{89.74\pm0.11}$
& $4.79\pm0.39$
& $6.81\pm0.52$
& $4.12\pm0.31$
& $5.24\pm0.32$ \\

nnU-Net-SHC$\times$2
& $90.08\pm0.29$
& $92.17\pm0.31$
& $86.59\pm0.27$
& $89.61\pm0.25$
& $4.61\pm0.34$
& $6.57\pm0.46$
& $4.21\pm0.29$
& $5.13\pm0.27$ \\

nnU-Net-SHC$\times$4
& $\mathbf{90.29\pm0.19}$
& $92.12\pm0.16$
& $86.79\pm0.18$
& $89.73\pm0.13$
& $\mathbf{4.36\pm0.27}$
& $\mathbf{6.34\pm0.43}$
& $\mathbf{3.74\pm0.25}$
& $\mathbf{4.81\pm0.24}$ \\

\midrule

VT-UNet
& $90.04\pm0.48$
& $91.88\pm0.24$
& $86.49\pm0.29$
& $89.47\pm0.22$
& $6.17\pm0.58$
& $6.52\pm0.61$
& $5.12\pm0.46$
& $5.94\pm0.38$ \\

VT-UNet-DHC$\times$2
& $\mathbf{90.63\pm0.13}$
& $\mathbf{91.96\pm0.16}$
& $\mathbf{86.57\pm0.14}$
& $\mathbf{89.72\pm0.11}$
& $\mathbf{4.71\pm0.35}$
& $\mathbf{5.69\pm0.41}$
& $\mathbf{3.91\pm0.29}$
& $\mathbf{4.77\pm0.25}$ \\

VT-UNet-DHC$\times$4
& $87.41\pm0.37$
& $90.81\pm0.28$
& $83.92\pm0.44$
& $86.05\pm0.31$
& $6.14\pm0.64$
& $7.71\pm0.73$
& $6.02\pm0.57$
& $6.62\pm0.46$ \\

VT-UNet-SHC$\times$2
& $88.37\pm0.39$
& $90.42\pm0.32$
& $85.34\pm0.36$
& $88.04\pm0.27$
& $7.69\pm0.72$
& $9.79\pm0.84$
& $4.26\pm0.38$
& $7.25\pm0.51$ \\

VT-UNet-SHC$\times$4
& $90.09\pm0.17$
& $91.64\pm0.19$
& $86.17\pm0.16$
& $89.30\pm0.14$
& $4.85\pm0.38$
& $7.32\pm0.59$
& $4.44\pm0.33$
& $5.54\pm0.36$ \\

\bottomrule
\end{tabular}%
}
\end{table}
\subsection{Modality reliance and clinical alignment}
\label{sec:dse}
We next examine whether HC changes sensitivity to the MRI sequence expected to be most informative for each tumor region. Table~\ref{tab:dse} reports the dominant-modality Dice drop $\Delta_{\text{dom}}$ and DSE (Eq.~\ref{eq:dse}) for every backbone--variant pair. Across the three backbones, the majority of HC configurations show a larger Dice drop than the baseline when the clinically correct sequence is removed, yielding positive DSE under both static and dynamic HC (T1ce for TC and ET, FLAIR for WT). The trend is clearest for whole-tumor (FLAIR) and tumor-core (T1ce) and weakest for enhancing tumor, where several DSE values are near zero, and a small number of configurations show no change or a slight decrease. On balance, adaptive connections tend to make these networks lean harder on the clinically appropriate evidence per region, which is exactly the region-conditioned reasoning a radiologist applies (T1ce for contrast-enhancing tissue; FLAIR for peritumoral edema), though the effect is not uniform across every variant. Fixed-connection baselines show no such sharpening, combining modalities the same way regardless of region. Because the effect appears in both the convolutional nnU-Net and the transformer-based SwinUNETR and VT-UNet, it stems from the adaptive connection rather than a single backbone. The full per-region, per-config modality-ablation profiles for all three backbones are provided in Appendix~\ref{app:ablation} (Figures~\ref{fig:modality_nnunet}, \ref{fig:modality_swin}, and~\ref{fig:modality_vt}).

\begin{table}[t]
\centering
\caption{Dominant-modality Dice drop $\Delta_{\text{dom}}$ and Dominant
Sensitivity Enhancement (DSE) for all backbone--variant pairs. Dominant sequence:
T1ce for TC and ET; FLAIR for WT. DSE is relative to the baseline
(Eq.~\ref{eq:dse}). Positive DSE (bold) indicates increased reliance on the clinically dominant sequence relative to baseline. Values near zero denote no meaningful change.}
\label{tab:dse}
\setlength{\tabcolsep}{5pt}
\resizebox{\linewidth}{!}{%
\begin{tabular}{llccccccccc}
\toprule
& & Baseline & \multicolumn{2}{c}{SHC$\times$2} & \multicolumn{2}{c}{SHC$\times$4}
  & \multicolumn{2}{c}{DHC$\times$2} & \multicolumn{2}{c}{DHC$\times$4} \\
\cmidrule(lr){3-3}\cmidrule(lr){4-5}\cmidrule(lr){6-7}\cmidrule(lr){8-9}\cmidrule(lr){10-11}
Model & Region & $\Delta_{\text{dom}}$ & $\Delta_{\text{dom}}$ & DSE & $\Delta_{\text{dom}}$ & DSE
      & $\Delta_{\text{dom}}$ & DSE & $\Delta_{\text{dom}}$ & DSE \\
\midrule
\multirow{3}{*}{nnU-Net}
 & TC & 0.43 & 0.50 & $+0.07$ & 0.39 & $-0.04$ & 0.61 & $+0.18$ & 0.43 & $+0.00$ \\
 & WT & 0.31 & 0.46 & $+0.15$ & 0.35 & $+0.04$ & 0.35 & $+0.04$ & 0.40 & $+0.09$ \\
 & ET & 0.80 & 0.75 & $-0.05$ & 0.68 & $-0.12$ & 0.84 & $+0.04$ & 0.81 & $+0.01$ \\
\midrule
\multirow{3}{*}{SwinUNETR}
 & TC & 0.50 & 0.49 & $-0.00$ & 0.51 & $+0.01$ & 0.50 & $+0.01$ & 0.46 & $-0.04$ \\
 & WT & 0.10 & 0.12 & $+0.02$ & 0.10 & $+0.01$ & 0.13 & $+0.03$ & 0.12 & $+0.02$ \\
 & ET & 0.88 & 0.89 & $+0.01$ & 0.89 & $+0.00$ & 0.90 & $+0.01$ & 0.89 & $+0.01$ \\
\midrule
\multirow{3}{*}{VT-UNet}
 & TC & 0.62 & 0.66 & $+0.04$ & 0.64 & $+0.02$ & 0.67 & $+0.05$ & 0.72 & $+0.10$ \\
 & WT & 0.25 & 0.34 & $+0.09$ & 0.31 & $+0.05$ & 0.35 & $+0.09$ & 0.34 & $+0.09$ \\
 & ET & 0.84 & 0.84 & $+0.00$ & 0.84 & $-0.00$ & 0.85 & $+0.01$ & 0.84 & $+0.01$ \\
\bottomrule
\end{tabular}%
}
\end{table}

\subsection{Where the reasoning happens: gate trajectories}
Figure~\ref{supp:beta_nnUNet3D} plots the depth gate $\beta$ per stream across the
encoder--bottleneck--decoder blocks of nnU-Net. Under DHC the streams swing well
above and below $\beta=1$ and diverge from each other, especially in the early
encoder and late decoder, showing the gate actively adapting at each depth. Under
SHC the streams settle into smoother, more separated trajectories that still
shift across depth but no longer oscillate sharply within a block. Across both
variants $\beta$ moves furthest from $1$ near the encoder and decoder ends and
stays closer to $1$ near the bottleneck, indicating that the model transforms
features most aggressively at shallow and deep layers while passing bottleneck
features through more conservatively. This localizes where adaptive feature integration is exercised across depth, complementing the modality-ablation analysis; it does not, on its own, establish that these gates encode clinically meaningful evidence selection. The full per-configuration $\beta$-evolution atlases for all three backbones are provided in Appendix~\ref{app:beta}.

\begin{table}[t]
\centering
\caption{Parameters added by HC (millions). $\Delta$ Rate (\%) is the increase
over the matching baseline. Any change in reasoning behavior cannot be attributed
to capacity: DHC adds $\leq 0.53\%$ and SHC $< 0.002\%$.}
\label{tab:params}
\resizebox{\linewidth}{!}{%
\begin{tabular}{lccc}
\toprule
Method & HC Params (M) & Total Params (M) & $\Delta$ Rate (\%) \\
\midrule
SwinUNETR-Base        & --       & 21.3406 & -- \\
SwinUNETR-DHC$\times$2 & 0.0174  & 21.3580 & $+0.08172$ \\
SwinUNETR-DHC$\times$4 & 0.0235  & 21.3640 & $+0.10991$ \\
SwinUNETR-SHC$\times$2 & 0.000128 & 21.3407 & $+0.00060$ \\
SwinUNETR-SHC$\times$4 & 0.000384 & 21.3410 & $+0.00180$ \\
\midrule
nnU-Net-Base        & --        & 30.3987 & -- \\
nnU-Net-DHC$\times$2 & 0.00678  & 30.4055 & $+0.02230$ \\
nnU-Net-DHC$\times$4 & 0.00912  & 30.4078 & $+0.02999$ \\
nnU-Net-SHC$\times$2 & 0.000048 & 30.3988 & $+0.00016$ \\
nnU-Net-SHC$\times$4 & 0.000144 & 30.3988 & $+0.00047$ \\
\midrule
VT-UNet-Base        & --        & 20.7571 & -- \\
VT-UNet-DHC$\times$2 & 0.04251  & 20.7996 & $+0.20478$ \\
VT-UNet-DHC$\times$4 & 0.11054  & 20.8676 & $+0.53253$ \\
VT-UNet-SHC$\times$2 & 0.000112 & 20.7572 & $+0.00054$ \\
VT-UNet-SHC$\times$4 & 0.000336 & 20.7574 & $+0.00162$ \\
\bottomrule
\end{tabular}%
}
\end{table}

\section{Discussion}

\textbf{Quantifying clinically grounded evidence reliance.}
Our analysis provides a quantitative and interpretable way to examine whether a
segmentation model depends on imaging information that is consistent with
clinical expectations. The modality-ablation profiles characterize how strongly
predictions for each tumor sub-region depend on individual MRI sequences, while
DSE measures how this dependence changes relative to the corresponding baseline.
The $\beta$ trajectories provide a complementary view of how adaptive feature
integration varies across network depth. Together, these analyses move beyond
segmentation accuracy alone and offer a structured way to study
\emph{modality-specific evidence reliance} in multimodal medical imaging models.

An important observation is that the clinically aligned sensitivity patterns
emerge without explicit supervision on modality importance. The training
objective does not encourage the model to preferentially rely on T1ce for TC or
ET, or on FLAIR for WT. Nevertheless, several HC configurations show increased
sensitivity to these clinically expected dominant modalities. This suggests that
adaptive feature integration can influence how multimodal evidence is represented
and combined, and may encourage modality-use patterns that are more consistent
with radiological expectations. At the same time, the effect is not uniform
across all architectures and configurations, indicating that adaptive
connections should be interpreted as a mechanism that can facilitate, rather
than guarantee, clinically aligned evidence reliance.

\textbf{Relevance to medical foundation models.} Although we study volumetric
segmentation, the probe is model-agnostic and modality-level, and applies
directly to multi-modal medical foundation and vision--language models: one can
ablate an input stream (an imaging sequence, a report, a pathology view) and read
out per-decision reliance and its clinical alignment via DSE. This offers the
reasoning community a lightweight, supervision-free interpretability tool for
checking whether a multi-modal medical model grounds its outputs in the correct
evidence, complementary to text-based chain-of-thought analysis.

\textbf{Limitations.} Our study is on a single benchmark (BraTS~2021) and does not
include a language modality or an explicit chain-of-thought interface; the
``reasoning'' we probe is the model's implicit evidence-selection rather than a
generated rationale. DSE depends on the ablation-by-zeroing operationalization and
on a fixed clinical prior for the dominant sequence per region. Extending the
probe to models that emit natural-language rationales, and to cross-modality
(image + text) fusion, is a natural next step.

\section{Conclusion}
We reframed multimodal brain tumor segmentation as a question of clinically grounded multimodal reasoning and introduced a supervision-free probe (modality ablation, the DSE metric, and gate-trajectory analysis) that shows which MRI sequence a model relies on for each tumor sub-region.
Replacing fixed connections with adaptive Hyper-Connections increased reliance on the clinically expected sequence (T1ce for TC and ET, FLAIR for WT) in most configurations across all three backbones, without any supervision on modality importance. The effect was clearest for WT and TC and weakest for ET.
Hyper-Connections also improved accuracy at very small cost, adding at most 0.53\% parameters. On validation, the best HC configuration raised mean Dice by 0.43 (nnU-Net), 0.81 (SwinUNETR), and 0.81 (VT-UNet) percentage points. On the held-out test set, boundary accuracy improved most for the weaker baselines: mean HD95 fell from 6.57 to 4.79\,mm for nnU-Net and from 5.88 to 4.71\,mm for VT-UNet. Gains were small or absent for the already strong SwinUNETR, and two VT-UNet configurations performed worse on the single-seed test split, so HC should be seen as a mechanism that often helps rather than one that always does.
Beyond these accuracy gains, our main contribution is a concrete, inspectable notion of clinical grounding for volumetric models. Because the probe works at the level of input modalities and needs no extra labels, it can be applied to other multimodal medical models, including medical foundation models.
medical foundation models.

\bibliography{ref}

@article{taha2015metrics,
  title={Metrics for evaluating 3D medical image segmentation: analysis, selection, and tool},
  author={Taha, Abdel Aziz and Hanbury, Allan},
  journal={BMC medical imaging},
  volume={15},
  number={1},
  pages={29},
  year={2015},
  publisher={Springer}
}

@article{wen2010updated,
  title={Updated response assessment criteria for high-grade gliomas: response assessment in neuro-oncology working group},
  author  = {Wen, Patrick Y. and Macdonald, David R. and Reardon, David A. and Cloughesy, Timothy F. and Sorensen, A. Gregory and Galanis, Evanthia and others},
  journal={Journal of clinical oncology},
  volume={28},
  number={11},
  pages={1963--1972},
  year={2010},
  publisher={American Society of Clinical Oncology}
}

@article{ellingson2015consensus,
    author  = {Ellingson, Benjamin M. and Bendszus, Martin and Boxerman, Jerrold and Barboriak, Daniel and Erickson, Bradley J. and Smits, Marion and others},
    title = {Consensus recommendations for a standardized brain tumor imaging protocol for clinical trials in brain metastases},
    journal = {Neuro-Oncology},
    volume = {22},
    number = {6},
    pages = {757-772},
    year = {2020},
    month = {06},
    issn = {1522-8517},
    doi = {10.1093/neuonc/noaa030},
    url = {https://doi.org/10.1093/neuonc/noaa030},
    eprint = {https://academic.oup.com/neuro-oncology/article-pdf/22/6/757/33373493/noaa030.pdf},
}

@article{menze2015brats,
  author  = {Menze, Bjoern H. and Jakab, Andras and Bauer, Stefan and Kalpathy-Cramer, Jayashree and Farahani, Keyvan and Kirby, Justin and others},
  journal={IEEE Transactions on Medical Imaging}, 
  title={The Multimodal Brain Tumor Image Segmentation Benchmark (BRATS)}, 
  year={2015},
  volume={34},
  number={10},
  pages={1993-2024},
  doi={10.1109/TMI.2014.2377694}}

@misc{baid2021rsnaasnrmiccaibrats2021benchmark,
      title={The RSNA-ASNR-MICCAI BraTS 2021 Benchmark on Brain Tumor Segmentation and Radiogenomic Classification}, 
      author={Ujjwal Baid and Satyam Ghodasara and Suyash Mohan and Michel Bilello and Evan Calabrese and Errol Colak and Keyvan Farahani and Jayashree Kalpathy-Cramer and Spyridon Bakas and others},
      year={2021},
      eprint={2107.02314},
      archivePrefix={arXiv},
      primaryClass={cs.CV},
      url={https://arxiv.org/abs/2107.02314}, 
}

@misc{zhu2025hyperconnections,
      title={Hyper-Connections}, 
      author={Defa Zhu and Hongzhi Huang and Zihao Huang and Yutao Zeng and Yunyao Mao and Banggu Wu and Qiyang Min and Xun Zhou},
      year={2025},
      eprint={2409.19606},
      archivePrefix={arXiv},
      primaryClass={cs.LG},
      url={https://arxiv.org/abs/2409.19606}, 
}

@misc{he2015deepresiduallearningimage,
      title={Deep Residual Learning for Image Recognition}, 
      author={Kaiming He and Xiangyu Zhang and Shaoqing Ren and Jian Sun},
      year={2015},
      eprint={1512.03385},
      archivePrefix={arXiv},
      primaryClass={cs.CV},
      url={https://arxiv.org/abs/1512.03385}, 
}

@misc{ronneberger2015unetconvolutionalnetworksbiomedical,
      title={U-Net: Convolutional Networks for Biomedical Image Segmentation}, 
      author={Olaf Ronneberger and Philipp Fischer and Thomas Brox},
      year={2015},
      eprint={1505.04597},
      archivePrefix={arXiv},
      primaryClass={cs.CV},
      url={https://arxiv.org/abs/1505.04597}, 
}

@article{isensee2021nnunet,
      title={nnU-Net: a self-configuring method for deep learning-based biomedical image segmentation},
      author={Fabian Isensee and Paul F. Jaeger and Simon A. A. Kohl and Jens Petersen and Klaus H. Maier-Hein},
      journal={Nature Methods},
      year={2021},
      volume={18},
      number={2},
      pages={203--211},
      doi={10.1038/s41592-020-01008-z},
      url={https://doi.org/10.1038/s41592-020-01008-z}
}

@misc{hatamizadeh2022swinunetrswintransformers,
      title={Swin UNETR: Swin Transformers for Semantic Segmentation of Brain Tumors in MRI Images}, 
      author={Ali Hatamizadeh and Vishwesh Nath and Yucheng Tang and Dong Yang and Holger Roth and Daguang Xu},
      year={2022},
      eprint={2201.01266},
      archivePrefix={arXiv},
      primaryClass={eess.IV},
      url={https://arxiv.org/abs/2201.01266}, 
}

@misc{peiris2022robustvolumetrictransformeraccurate,
      title={A Robust Volumetric Transformer for Accurate 3D Tumor Segmentation}, 
      author={Himashi Peiris and Munawar Hayat and Zhaolin Chen and Gary Egan and Mehrtash Harandi},
      year={2022},
      eprint={2111.13300},
      archivePrefix={arXiv},
      primaryClass={eess.IV},
      url={https://arxiv.org/abs/2111.13300}, 
}

@misc{bachlechner2020rezeroneedfastconvergence,
      title={ReZero is All You Need: Fast Convergence at Large Depth}, 
      author={Thomas Bachlechner and Bodhisattwa Prasad Majumder and Huanru Henry Mao and Garrison W. Cottrell and Julian McAuley},
      year={2020},
      eprint={2003.04887},
      archivePrefix={arXiv},
      primaryClass={cs.LG},
      url={https://arxiv.org/abs/2003.04887}, 
}

@misc{xie2023residualtransformerdualresidual,
      title={ResiDual: Transformer with Dual Residual Connections}, 
      author={Shufang Xie and Huishuai Zhang and Junliang Guo and Xu Tan and Jiang Bian and Hany Hassan Awadalla and Arul Menezes and Tao Qin and Rui Yan},
      year={2023},
      eprint={2304.14802},
      archivePrefix={arXiv},
      primaryClass={cs.CL},
      url={https://arxiv.org/abs/2304.14802}, 
}

@misc{loshchilov2019decoupledweightdecayregularization,
      title={Decoupled Weight Decay Regularization}, 
      author={Ilya Loshchilov and Frank Hutter},
      year={2019},
      eprint={1711.05101},
      archivePrefix={arXiv},
      primaryClass={cs.LG},
      url={https://arxiv.org/abs/1711.05101}, 
}

@misc{oktay2018attentionunetlearninglook,
      title={Attention U-Net: Learning Where to Look for the Pancreas}, 
      author={Ozan Oktay and Jo Schlemper and Loic Le Folgoc and Matthew Lee and Mattias Heinrich and Kazunari Misawa and Kensaku Mori and Steven McDonagh and Nils Y Hammerla and Bernhard Kainz and Ben Glocker and Daniel Rueckert},
      year={2018},
      eprint={1804.03999},
      archivePrefix={arXiv},
      primaryClass={cs.CV},
      url={https://arxiv.org/abs/1804.03999}, 
}
\onecolumn
\appendix
\section{Modality-Ablation Profiles}
\label{app:ablation}

This appendix provides the complete modality-ablation results for nnU-Net,
SwinUNETR, and VT-UNet. For each trained model, we independently remove one of
the four MRI sequences T1, T1ce, T2, or FLAIR at inference time by setting
the corresponding input channel to zero. We then measure the change in Dice for
Tumor Core (TC), Whole Tumor (WT), and Enhancing Tumor (ET), following
Eq.~\eqref{eq:reliance}. A larger positive Dice decrease indicates greater
sensitivity of the corresponding sub-region prediction to the removed modality.
Conversely, a value close to zero indicates limited sensitivity, while a negative
value indicates that segmentation performance increases after the modality is
removed.

\begin{figure}[p]
  \centering
  \setlength{\tabcolsep}{3pt}
  \renewcommand{\arraystretch}{1}
  \begin{tabular}{@{}c c@{}}
    \rotatebox[origin=c]{90}{\footnotesize\textbf{Baseline}} &
    \raisebox{-0.5\height}{\includegraphics[height=0.180\textheight, width=1\linewidth, keepaspectratio, trim=0 10 0 30, clip]{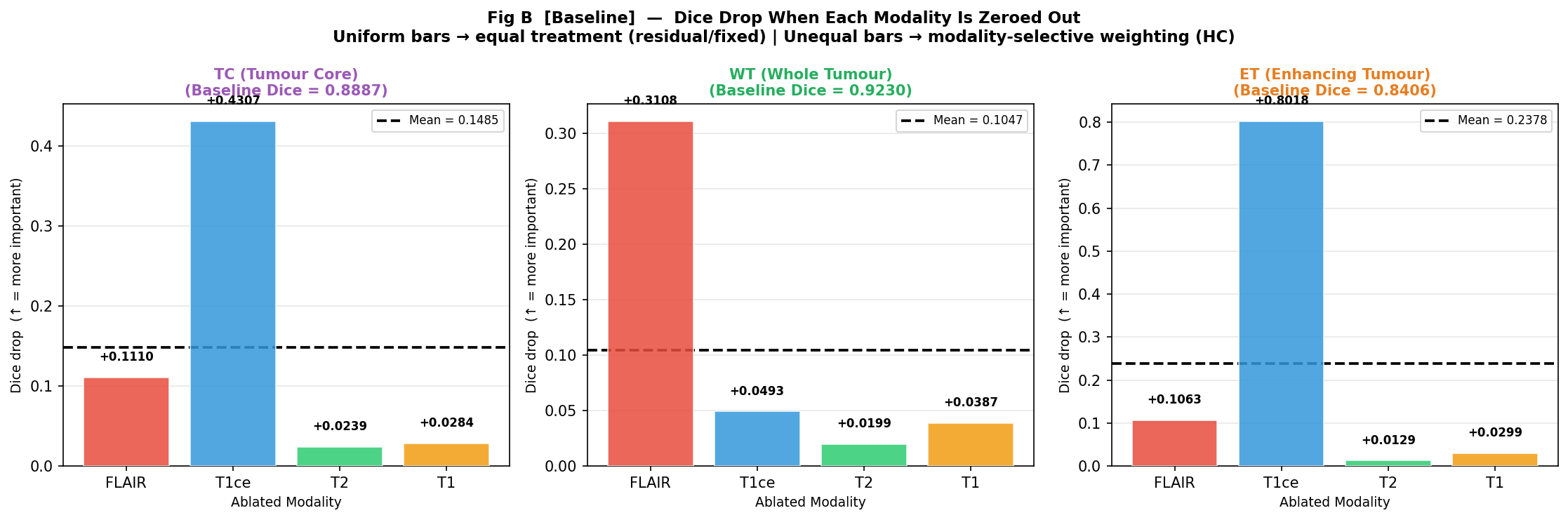}} \\[4pt]
    \rotatebox[origin=c]{90}{\footnotesize\textbf{SHC-$\times$2}} &
    \raisebox{-0.5\height}{\includegraphics[height=0.180\textheight, width=1\linewidth, keepaspectratio, trim=0 10 0 30, clip]{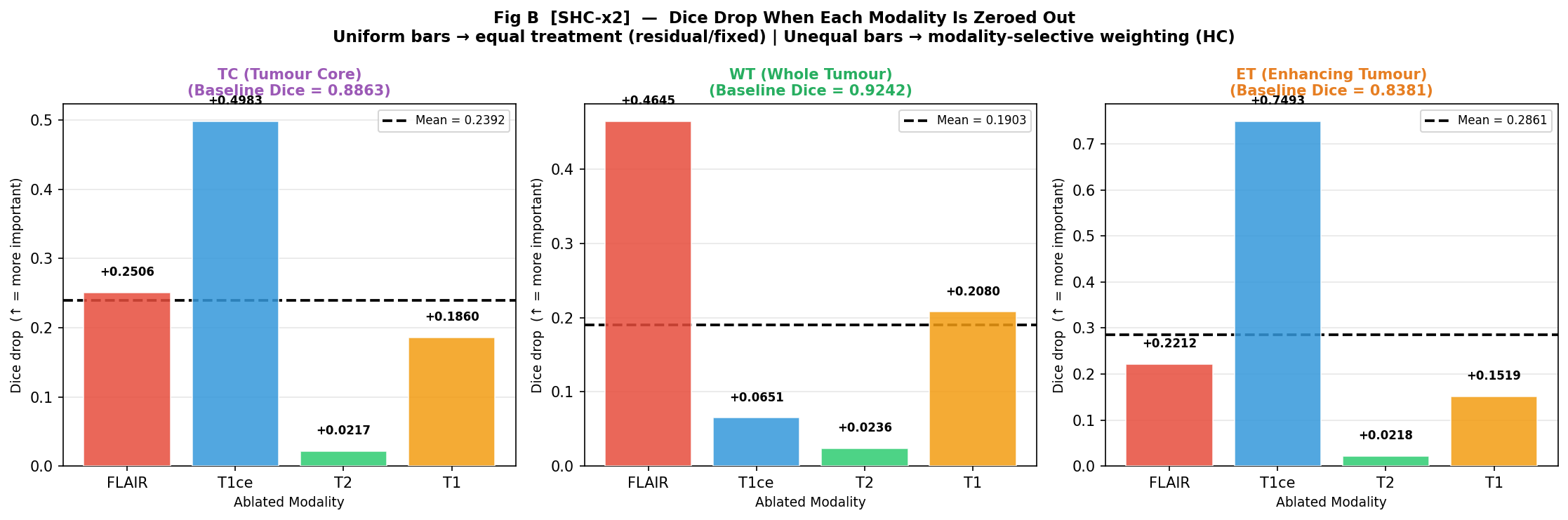}} \\[4pt]
    \rotatebox[origin=c]{90}{\footnotesize\textbf{SHC-$\times$4}} &
    \raisebox{-0.5\height}{\includegraphics[height=0.180\textheight, width=1\linewidth, keepaspectratio, trim=0 10 0 30, clip]{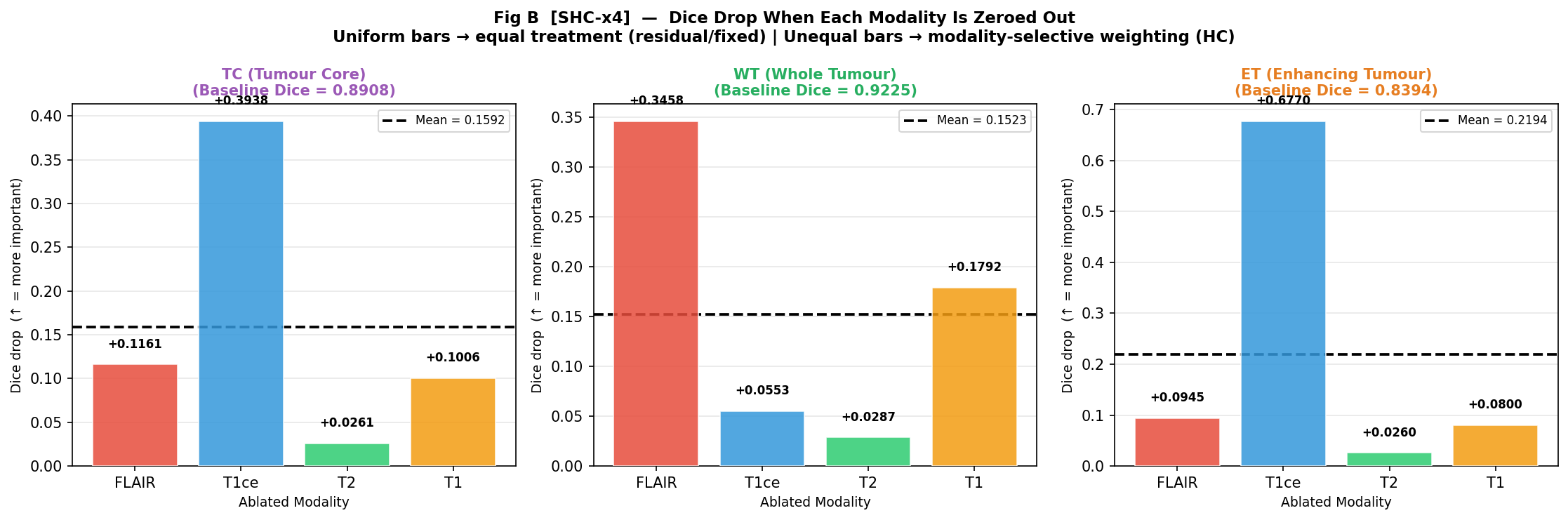}} \\[4pt]
    \rotatebox[origin=c]{90}{\footnotesize\textbf{DHC-$\times$2}} &
    \raisebox{-0.5\height}{\includegraphics[height=0.180\textheight, width=1\linewidth, keepaspectratio, trim=0 10 0 30, clip]{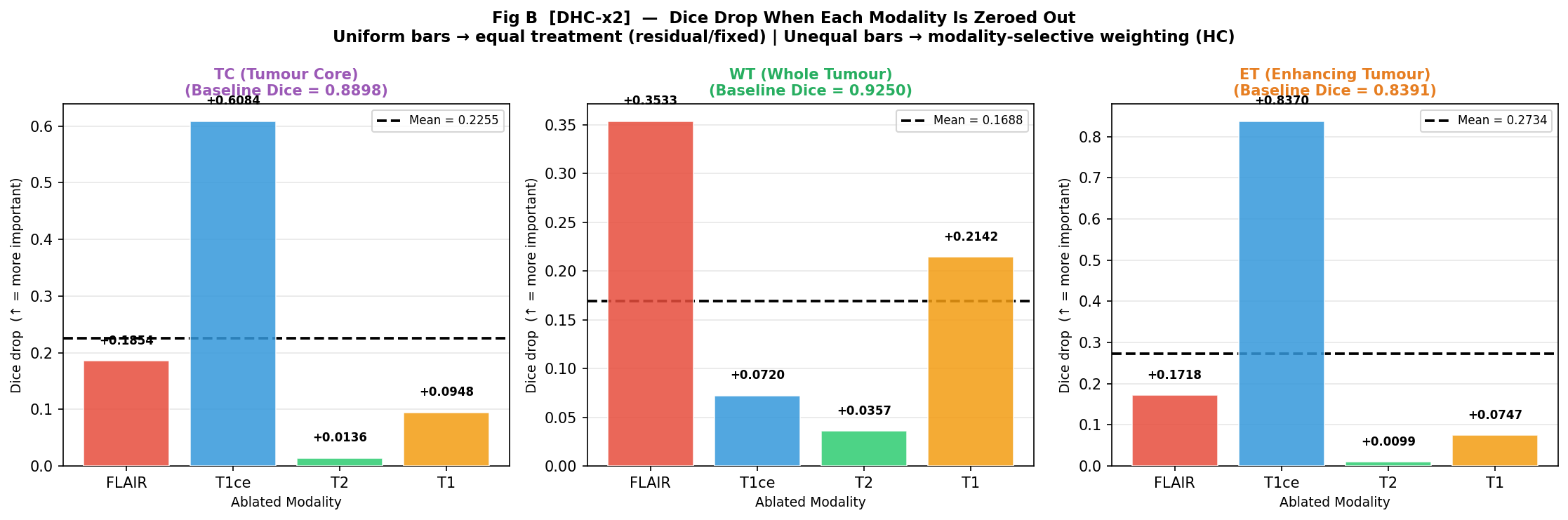}} \\[4pt]
    \rotatebox[origin=c]{90}{\footnotesize\textbf{DHC-$\times$4}} &
    \raisebox{-0.5\height}{\includegraphics[height=0.180\textheight, width=1\linewidth, keepaspectratio, trim=0 10 0 30, clip]{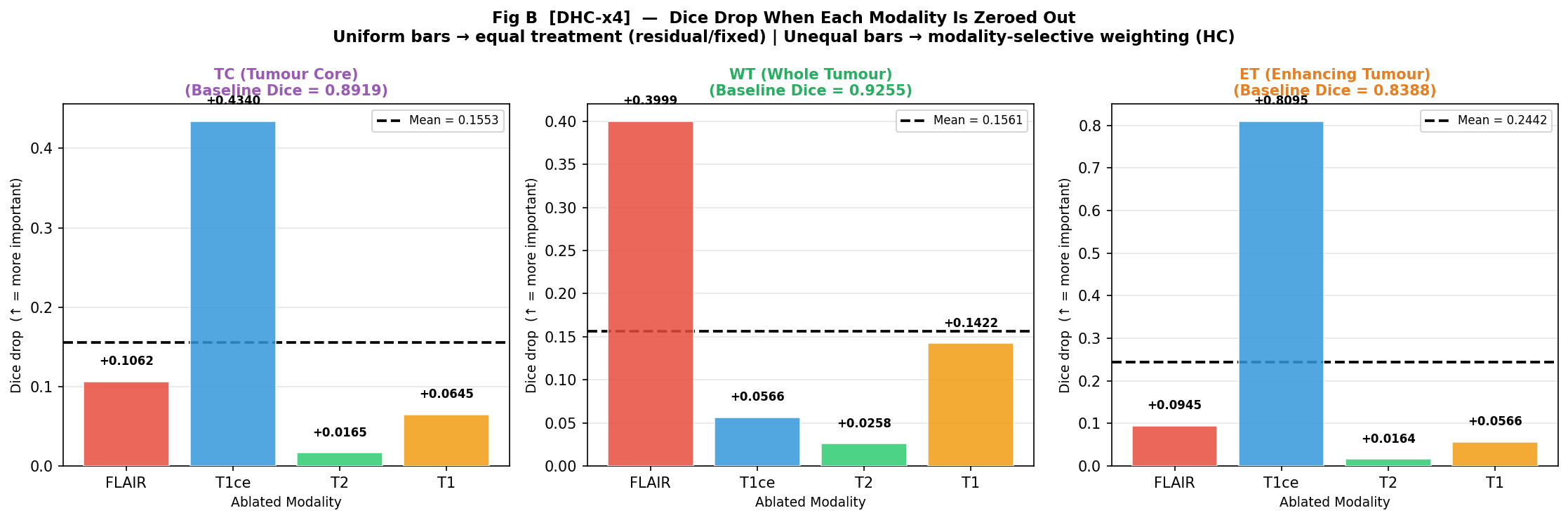}} \\
  \end{tabular}
  \caption{\textbf{Modality ablation (nnU-Net).} Dice drop when each MRI sequence is removed at inference, shown per configuration (rows: Baseline, SHC$\times$2/$\times$4, DHC$\times$2/$\times$4) and tumor region (columns: TC, WT, ET). The dashed line is the mean drop $\bar{\Delta}$. HC variants increase sensitivity to T1ce (TC/ET) and FLAIR (WT) relative to the baseline.}
  \label{fig:modality_nnunet}
\end{figure}

\begin{figure}[p]
  \centering
  \setlength{\tabcolsep}{3pt}
  \renewcommand{\arraystretch}{1}
  \begin{tabular}{@{}c c@{}}
    \rotatebox[origin=c]{90}{\footnotesize\textbf{Baseline}} &
    \raisebox{-0.5\height}{\includegraphics[height=0.180\textheight, width=1\linewidth, keepaspectratio, trim=0 10 0 30, clip]{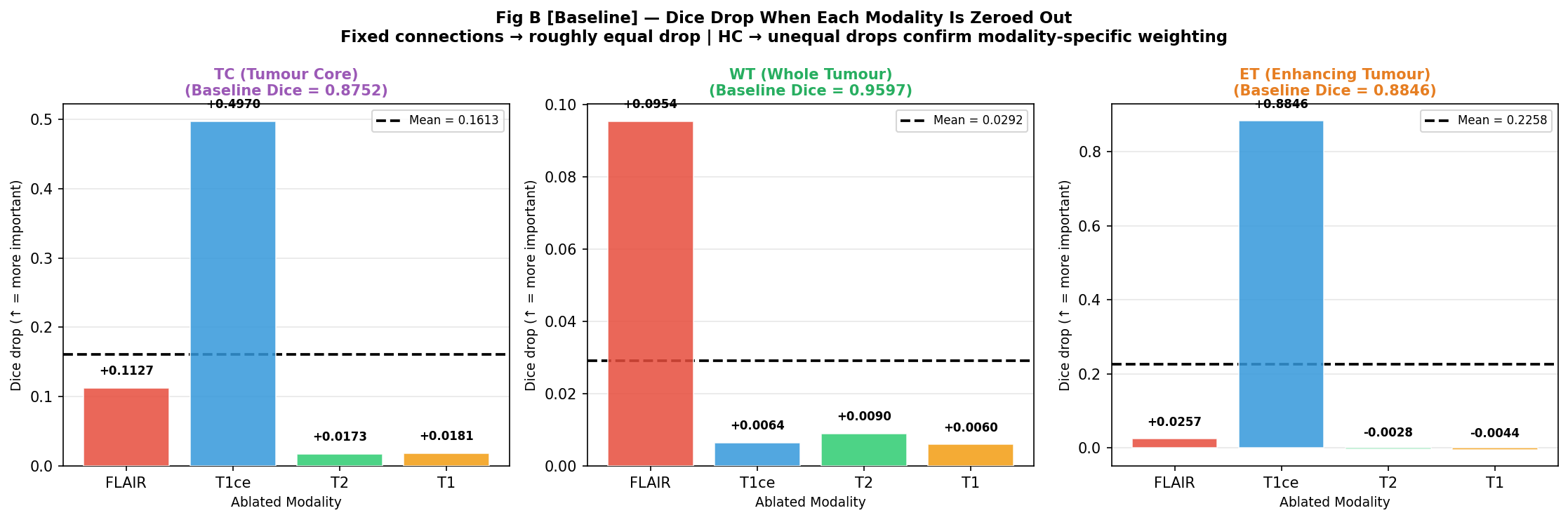}} \\[4pt]
    \rotatebox[origin=c]{90}{\footnotesize\textbf{SHC-$\times$2}} &
    \raisebox{-0.5\height}{\includegraphics[height=0.180\textheight, width=1\linewidth, keepaspectratio, trim=0 10 0 30, clip]{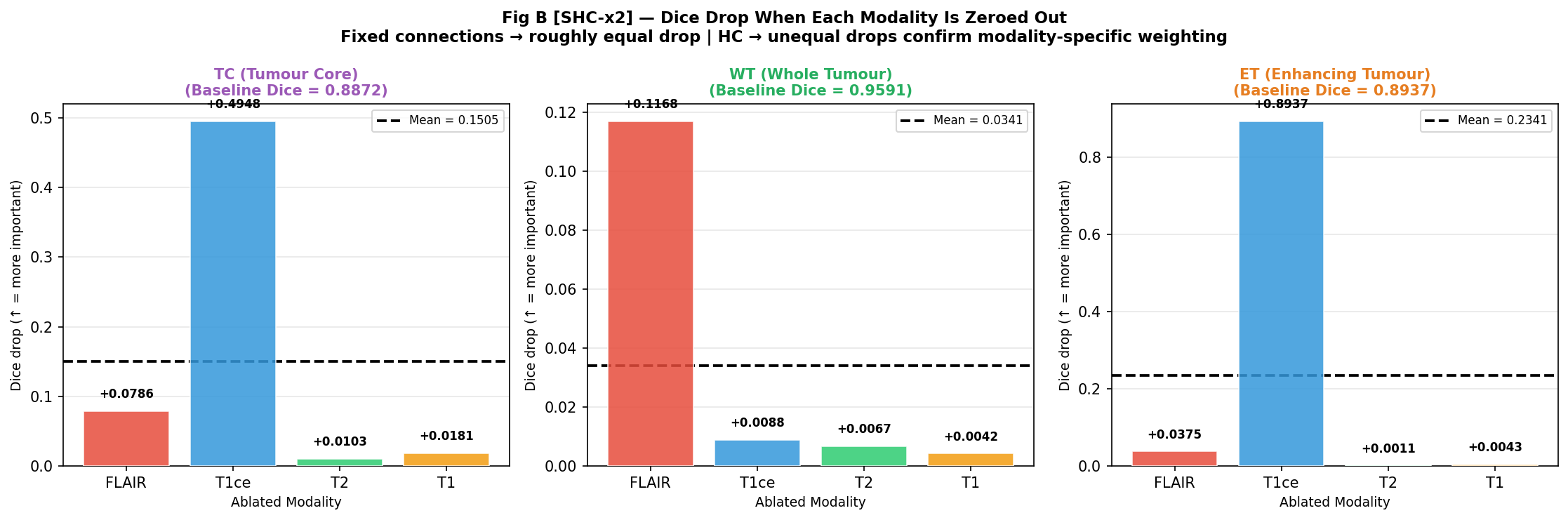}} \\[4pt]
    \rotatebox[origin=c]{90}{\footnotesize\textbf{SHC-$\times$4}} &
    \raisebox{-0.5\height}{\includegraphics[height=0.180\textheight, width=1\linewidth, keepaspectratio, trim=0 10 0 30, clip]{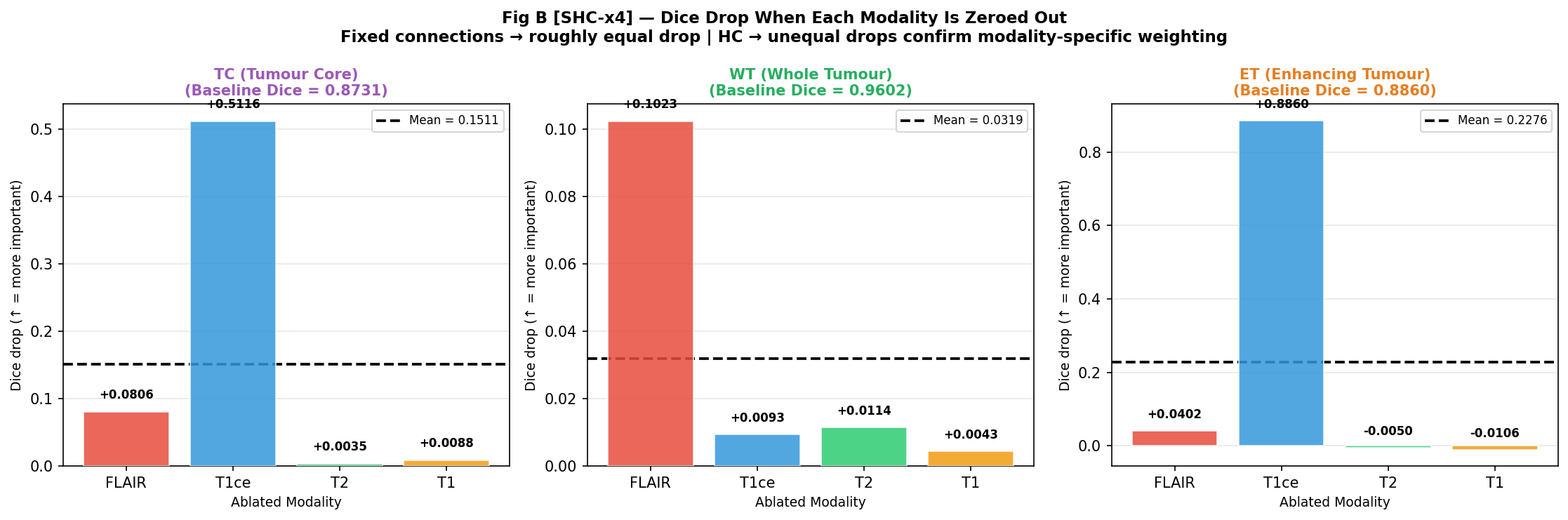}} \\[4pt]
    \rotatebox[origin=c]{90}{\footnotesize\textbf{DHC-$\times$2}} &
    \raisebox{-0.5\height}{\includegraphics[height=0.180\textheight, width=1\linewidth, keepaspectratio, trim=0 10 0 30, clip]{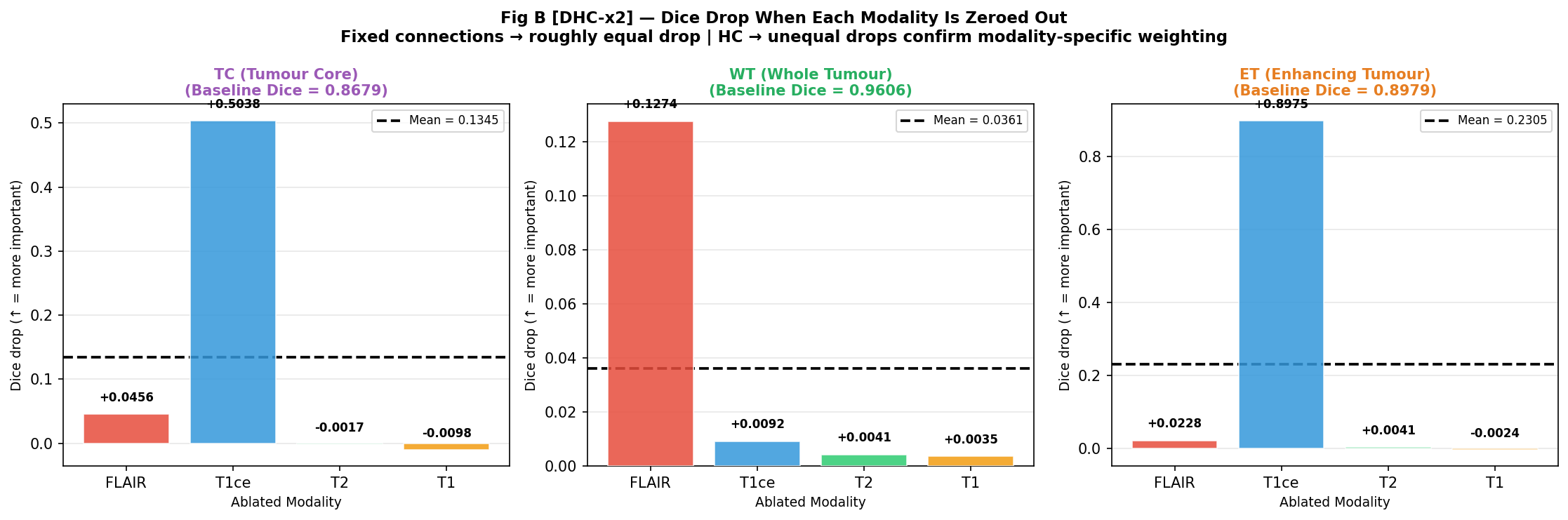}} \\[4pt]
    \rotatebox[origin=c]{90}{\footnotesize\textbf{DHC-$\times$4}} &
    \raisebox{-0.5\height}{\includegraphics[height=0.180\textheight, width=1\linewidth, keepaspectratio, trim=0 10 0 30, clip]{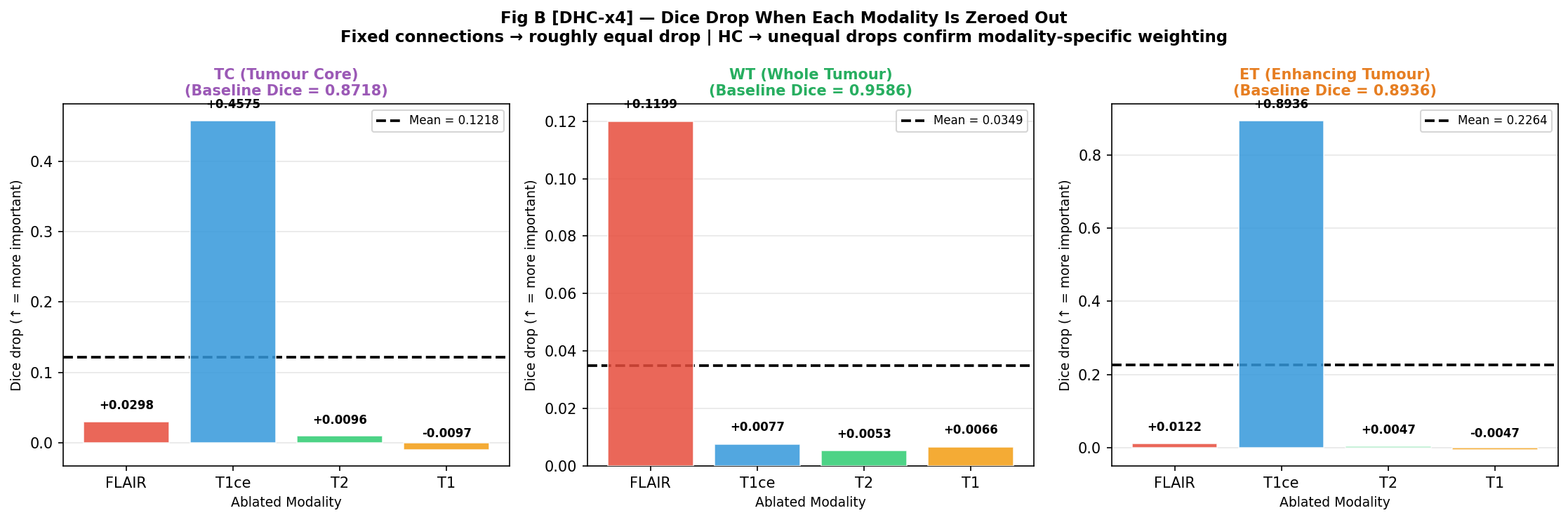}} \\
  \end{tabular}
\caption{\textbf{Modality ablation (SwinUNETR).} As in Figure~\ref{fig:modality_nnunet}, for SwinUNETR. HC variants increase sensitivity to T1ce (TC/ET) and FLAIR (WT).}
\label{fig:modality_swin}
\end{figure}

\begin{figure}[p]
  \centering
  \setlength{\tabcolsep}{3pt}
  \renewcommand{\arraystretch}{1}
  \begin{tabular}{@{}c c@{}}
    \rotatebox[origin=c]{90}{\footnotesize\textbf{Baseline}} &
    \raisebox{-0.5\height}{\includegraphics[height=0.180\textheight, width=1\linewidth, keepaspectratio, trim=0 10 0 30, clip]{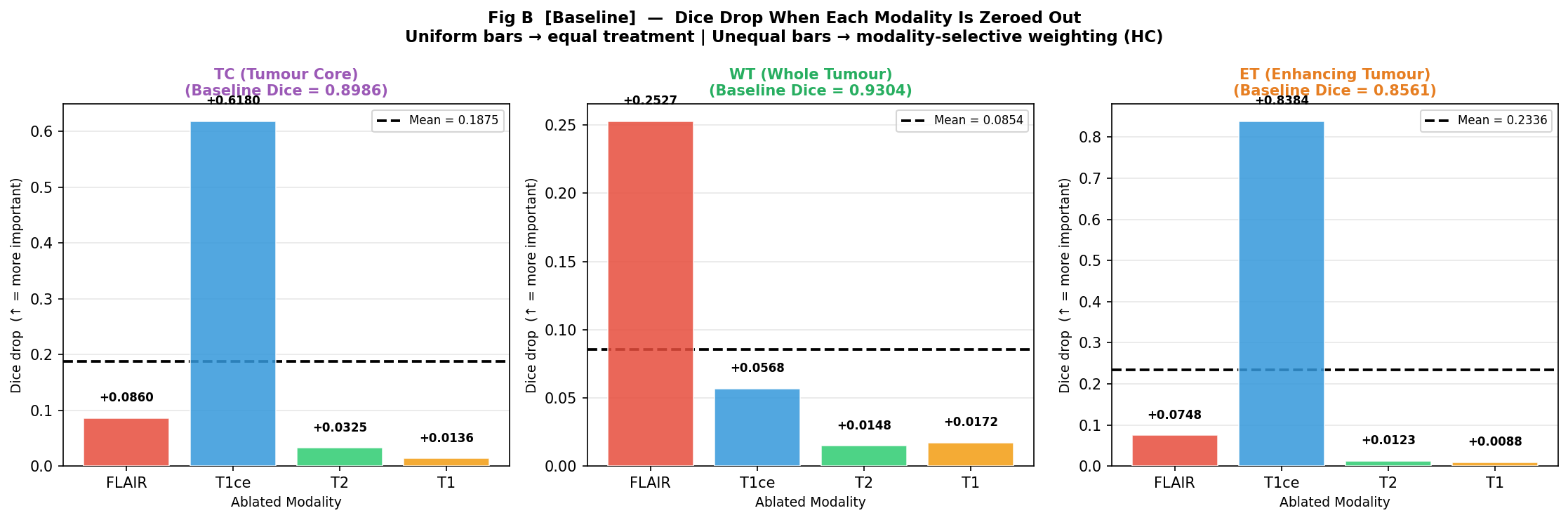}} \\[4pt]
    \rotatebox[origin=c]{90}{\footnotesize\textbf{SHC-$\times$2}} &
    \raisebox{-0.5\height}{\includegraphics[height=0.180\textheight, width=1\linewidth, keepaspectratio, trim=0 10 0 30, clip]{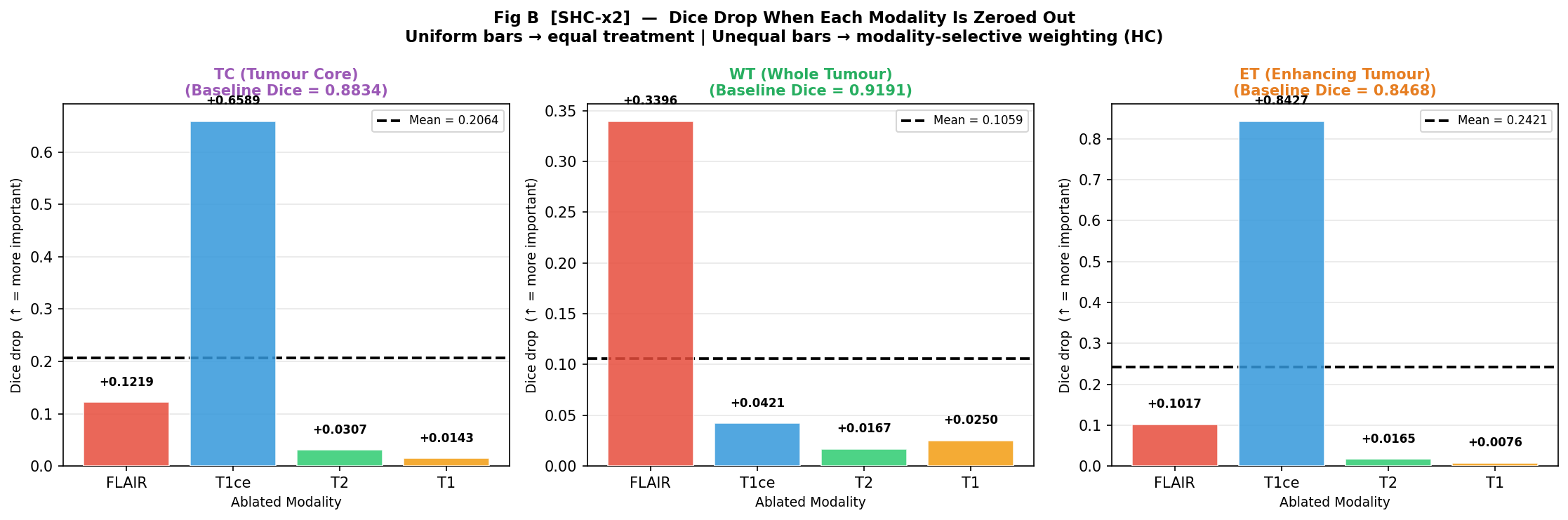}} \\[4pt]
    \rotatebox[origin=c]{90}{\footnotesize\textbf{SHC-$\times$4}} &
    \raisebox{-0.5\height}{\includegraphics[height=0.180\textheight, width=1\linewidth, keepaspectratio, trim=0 10 0 30, clip]{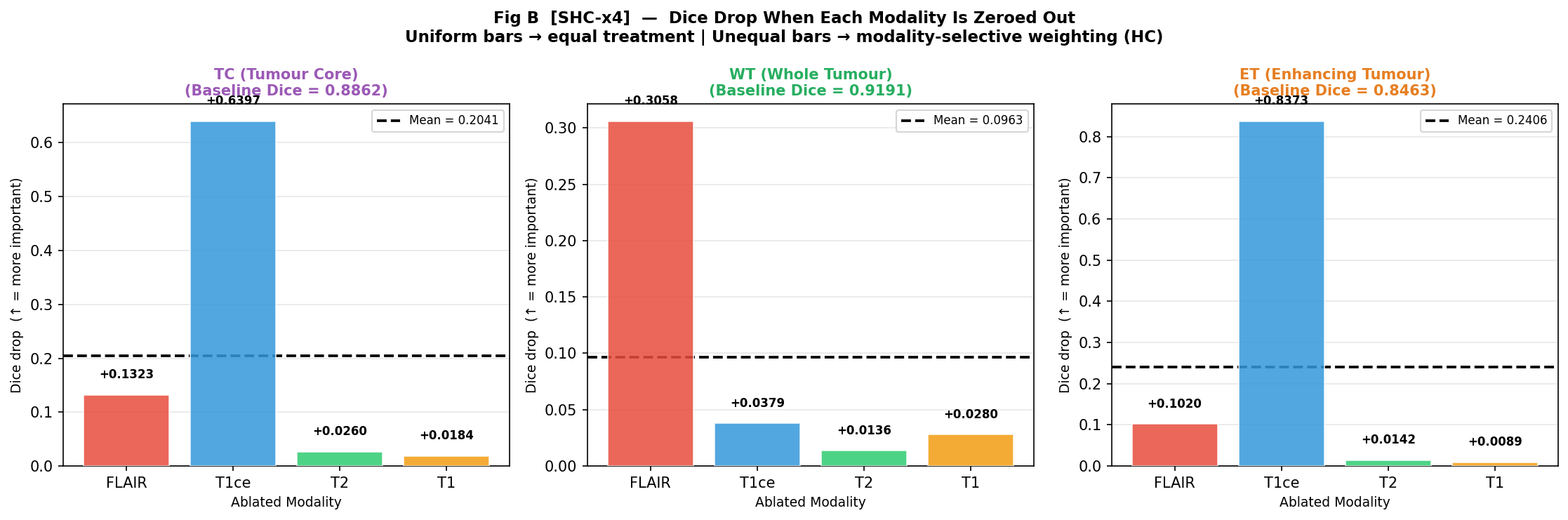}} \\[4pt]
    \rotatebox[origin=c]{90}{\footnotesize\textbf{DHC-$\times$2}} &
    \raisebox{-0.5\height}{\includegraphics[height=0.180\textheight, width=1\linewidth, keepaspectratio, trim=0 10 0 30, clip]{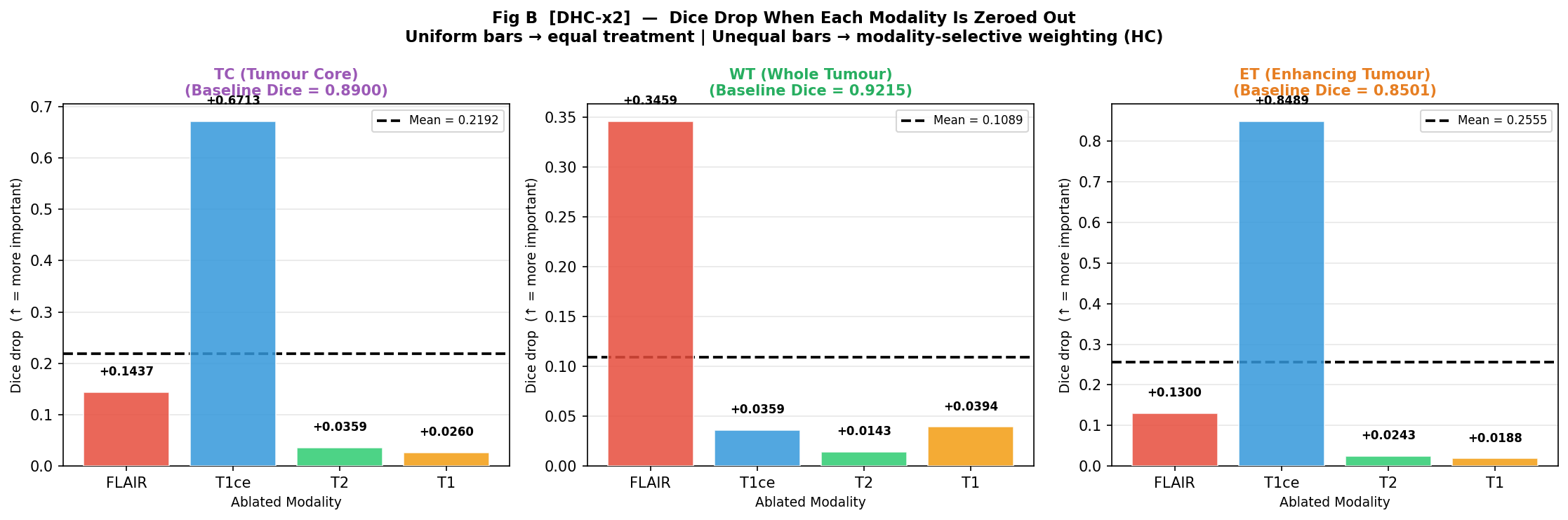}} \\[4pt]
    \rotatebox[origin=c]{90}{\footnotesize\textbf{DHC-$\times$4}} &
    \raisebox{-0.5\height}{\includegraphics[height=0.180\textheight, width=1\linewidth, keepaspectratio, trim=0 10 0 30, clip]{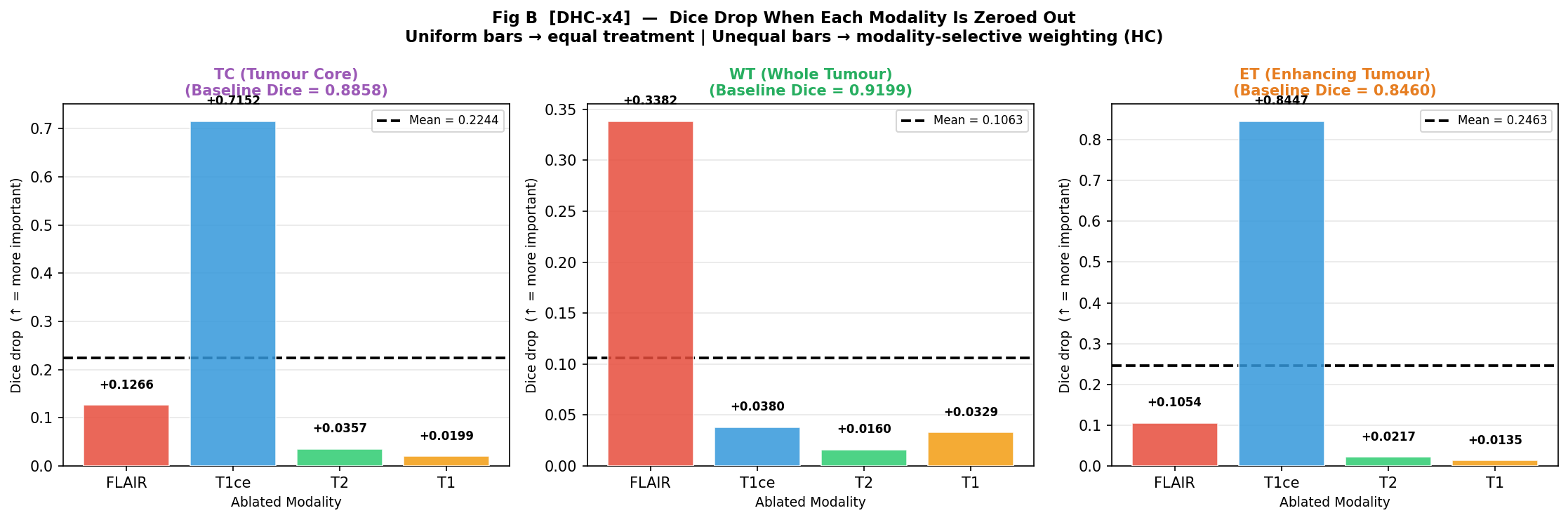}} \\
  \end{tabular}
\caption{\textbf{Modality ablation (VT-UNet).} As in Figure~\ref{fig:modality_nnunet}, for VT-UNet; the same T1ce (TC/ET) and FLAIR (WT) pattern holds, confirming the effect on transformer backbones.}
\label{fig:modality_vt}
\end{figure}

Figures~\ref{fig:modality_nnunet}--\ref{fig:modality_vt} report the complete
ablation profiles for the baseline, SHC$\times$2, SHC$\times$4,
DHC$\times$2, and DHC$\times$4 configurations. Each row corresponds to a
connection configuration, and the columns show modality sensitivity separately
for TC, WT, and ET. This presentation complements the DSE summary in
Table~\ref{tab:dse} by retaining the response to \emph{all four} input
modalities rather than considering only the clinically expected dominant
sequence.

\section{Qualitative Segmentation Results}

\label{app:qualitative}

\begin{figure}[p]
    \centering
    \includegraphics[width=\linewidth,height=0.92\textheight,keepaspectratio]{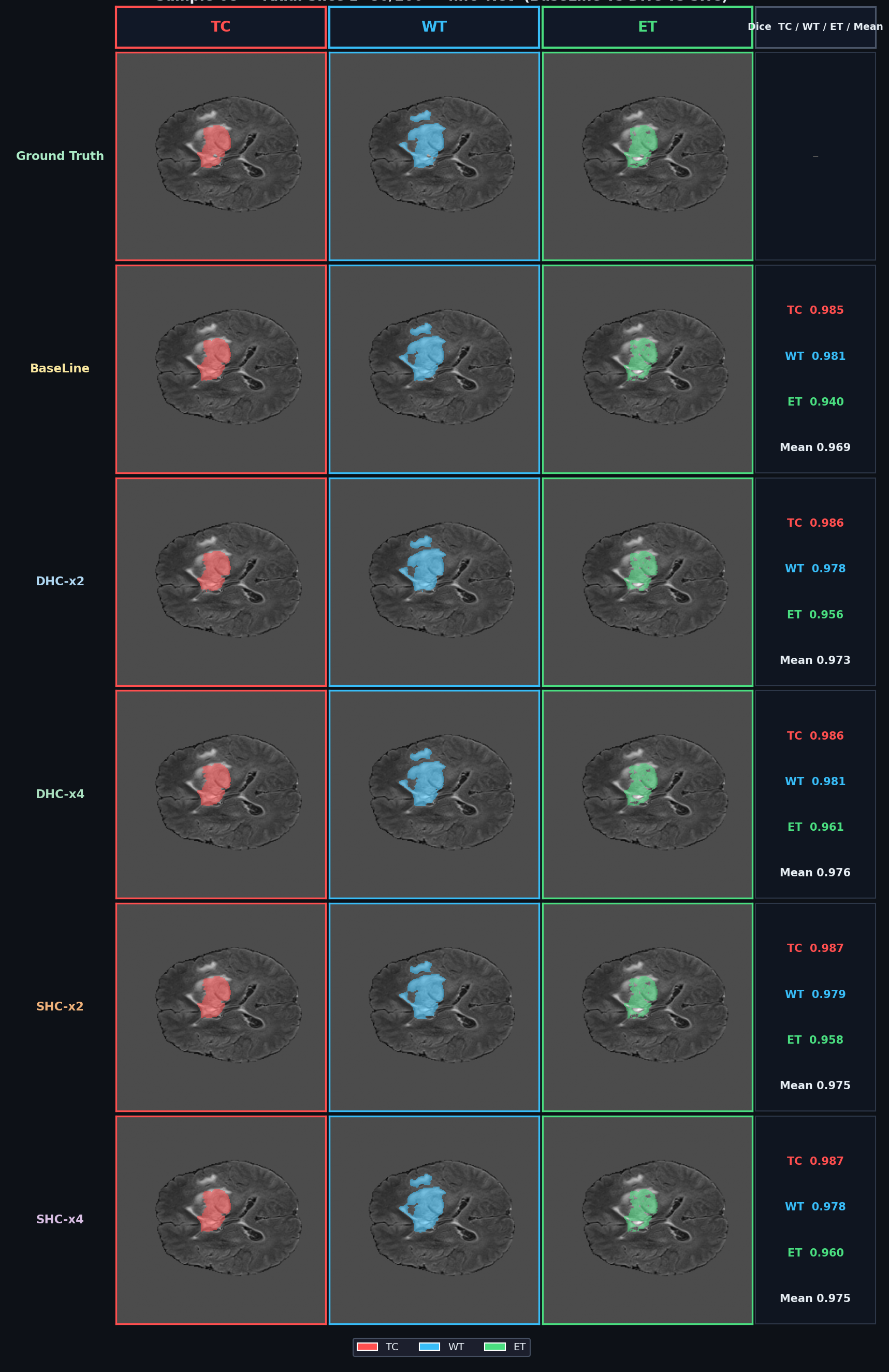}
    \caption{Qualitative results for nnU-Net.
    Rows: Ground Truth, Baseline, DHC$\times$2, DHC$\times$4, SHC$\times$2, SHC$\times$4.
    Columns: TC (red), WT (blue), ET (green). Dice scores per configuration are shown on the right.
    HC-based models match ground-truth boundaries more closely, particularly for ET.}
    \label{fig:nnunet_preds}
\end{figure}

\begin{figure}[p]
    \centering
    \includegraphics[width=\linewidth,height=0.92\textheight,keepaspectratio]{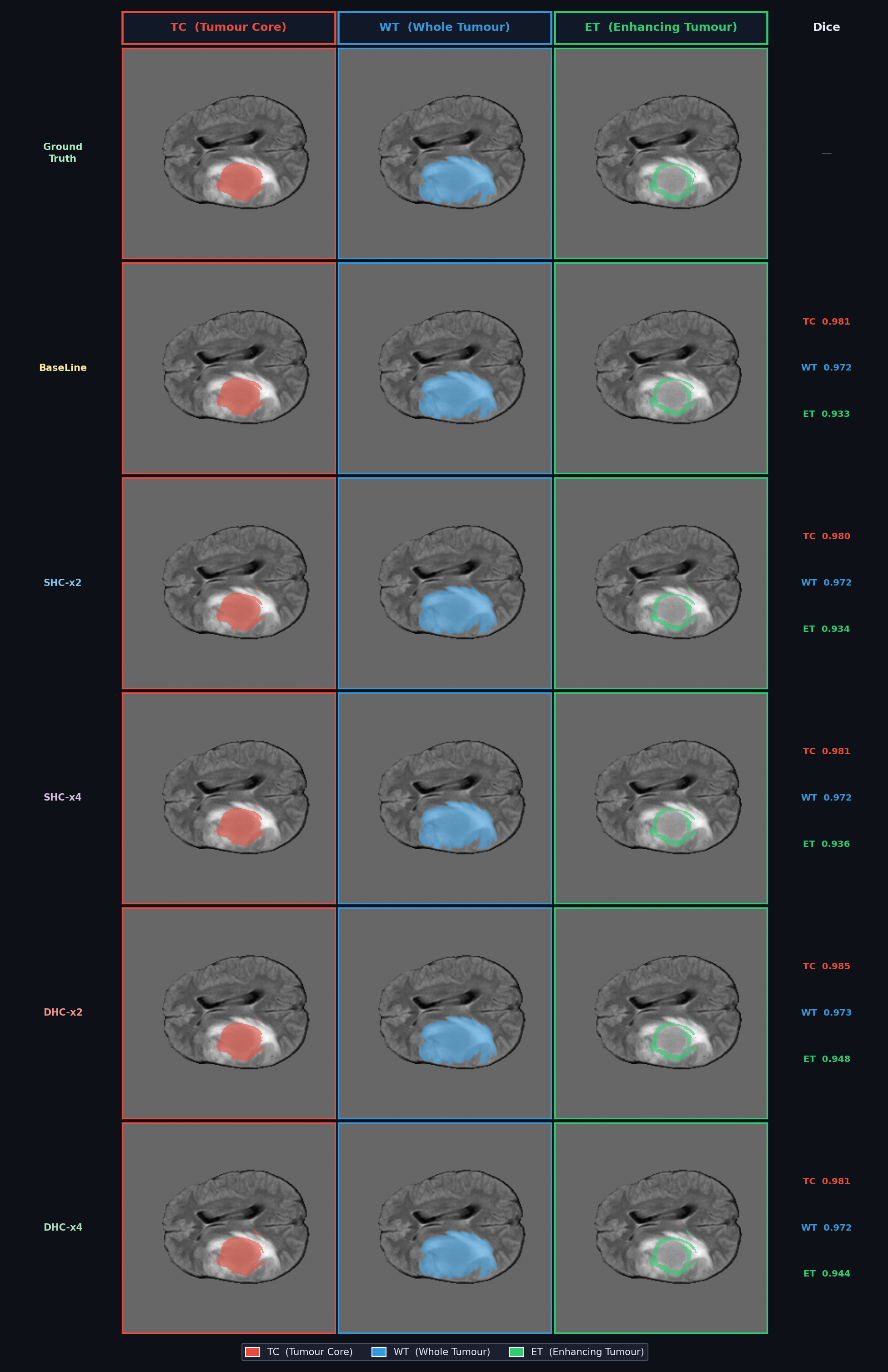}
    \caption{Qualitative results for SwinUNETR.}
    \label{fig:swin_preds}
\end{figure}

\begin{figure}[p]
    \centering
    \includegraphics[width=\linewidth,height=0.92\textheight,keepaspectratio]{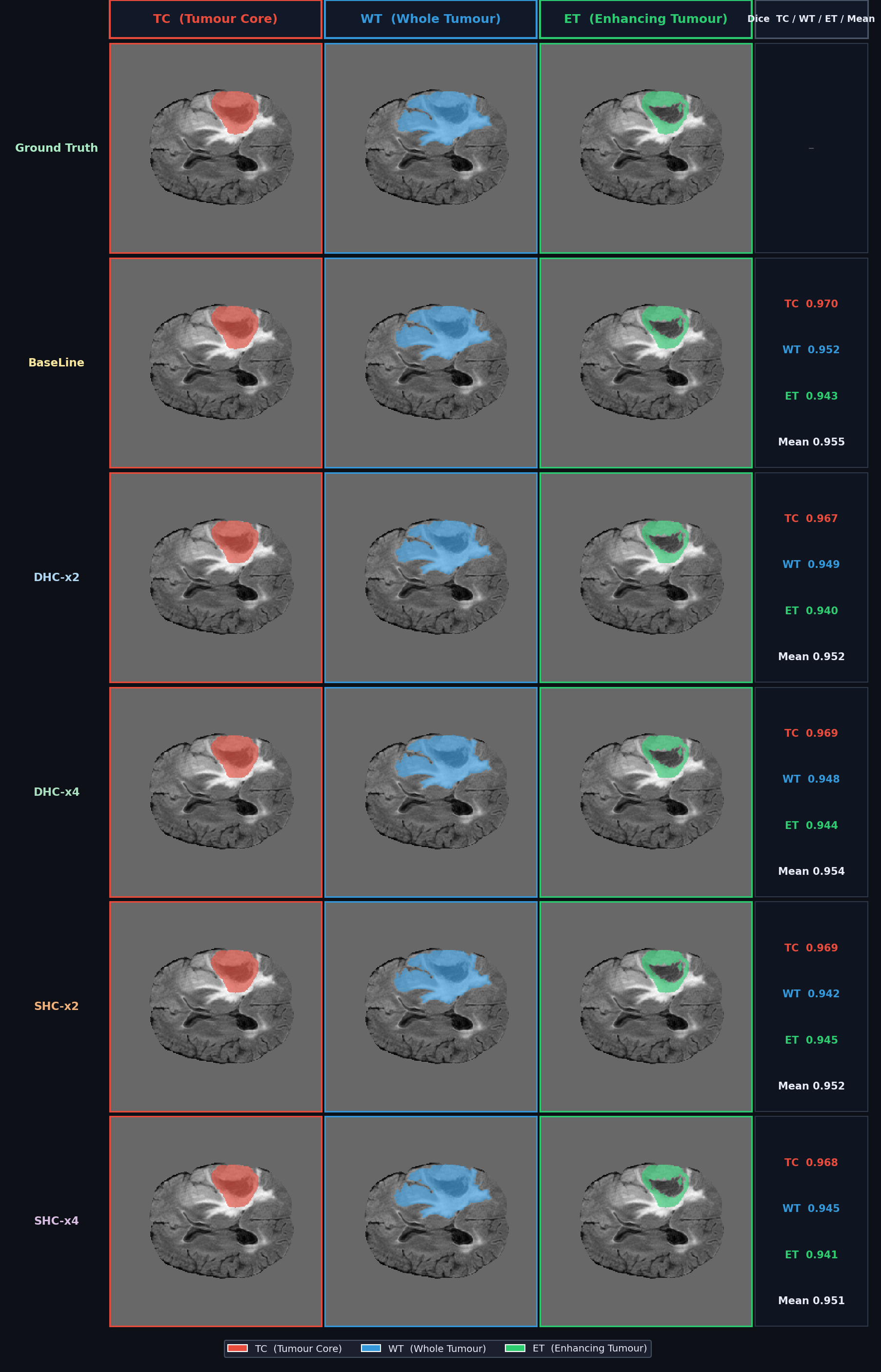}
    \caption{Qualitative results for VT-UNet.}
    \label{fig:vt_preds}
\end{figure}

This section provides predictions on all the models with qualitative comparisons between baseline and HC-based models. Figures \ref{fig:nnunet_preds}, \ref{fig:swin_preds}, and \ref{fig:vt_preds} provide qualitative examples from the held-out test set for nnU-Net, SwinUNETR, and VT-UNet. For each backbone, predictions from the baseline and the four HC configurations are shown
alongside the ground-truth segmentation. We visualize Tumor Core (TC), Whole
Tumor (WT), and Enhancing Tumor (ET) separately to facilitate comparison of
region-specific segmentation behavior.

The qualitative examples are intended to complement, rather than replace, the
quantitative Dice and HD95 results in Table~\ref{tab:test}. In several cases,
HC-based predictions show improved agreement with the annotated tumor extent,
particularly around small or irregular enhancing regions and along difficult
boundaries. Such differences are relevant because small boundary deviations may
have a modest effect on Dice while producing a larger change in HD95.

\section{Beta Evolution Analysis}
\label{app:beta}

We analyze the depth-wise gate $\beta$ to characterize how Hyper-Connections modify feature integration across the network. Under our parameterization, $\beta=1$ corresponds to the reference residual behavior, while deviations from $1$ indicate changes in the relative contribution of the transformed and skip pathways. Importantly, $\beta$ is not a direct measure of modality importance; rather, it provides an internal view of how feature mixing evolves across network depth.

Figures~\ref{supp:beta_nnUNet3D}--\ref{supp:beta_VTUNet3D} show the $\beta$ trajectories for DHC$\times$2, DHC$\times$4, SHC$\times$2, and SHC$\times$4. DHC generally exhibits greater variation across streams and network depth, consistent with its input-dependent mixing, whereas SHC produces smoother and more stable trajectories. These patterns support the modality-ablation analysis by showing \emph{where} adaptive feature integration changes within the network, without interpreting the gates themselves as direct evidence of clinical reasoning.




\newcommand{\betaAtlas}[2]{%
  \begin{figure}[p]
    \centering
    \includegraphics[width=\linewidth,height=0.23\textheight,trim=0 8 0 8,clip]{beta_plots_unified/beta_DHC-x2_#2.png}\\[2pt]
    \includegraphics[width=\linewidth,height=0.23\textheight,trim=0 8 0 8,clip]{beta_plots_unified/beta_DHC-x4_#2.png}\\[2pt]
    \includegraphics[width=\linewidth,height=0.23\textheight,trim=0 8 0 8,clip]{beta_plots_unified/beta_SHC-x2_#2.png}\\[2pt]
    \includegraphics[width=\linewidth,height=0.23\textheight,trim=0 8 0 8,clip]{beta_plots_unified/beta_SHC-x4_#2.png}
    \caption{%
      \textbf{Beta evolution atlas #1.}
      Top to bottom: DHC-$\times$2, DHC-$\times$4, SHC-$\times$2, SHC-$\times$4.
      A rising $\beta$ indicates increasing reliance on hybrid connections,
      while a flatter or decreasing trend suggests dominance of skip pathways.%
    }
    \label{supp:beta_#2}
  \end{figure}
}

\betaAtlas{nnUNet3D}{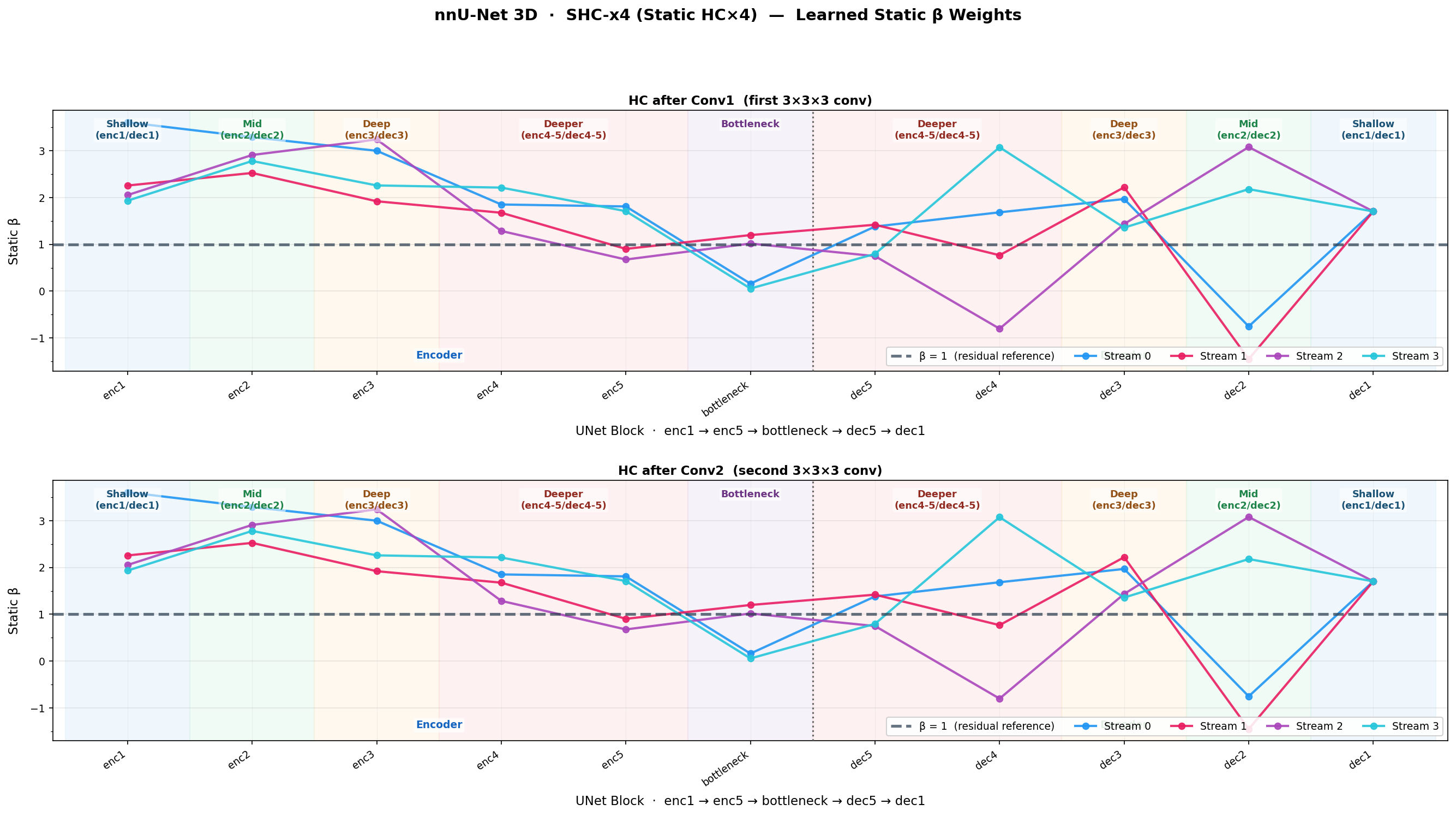}
\betaAtlas{SwinUNETR3D}{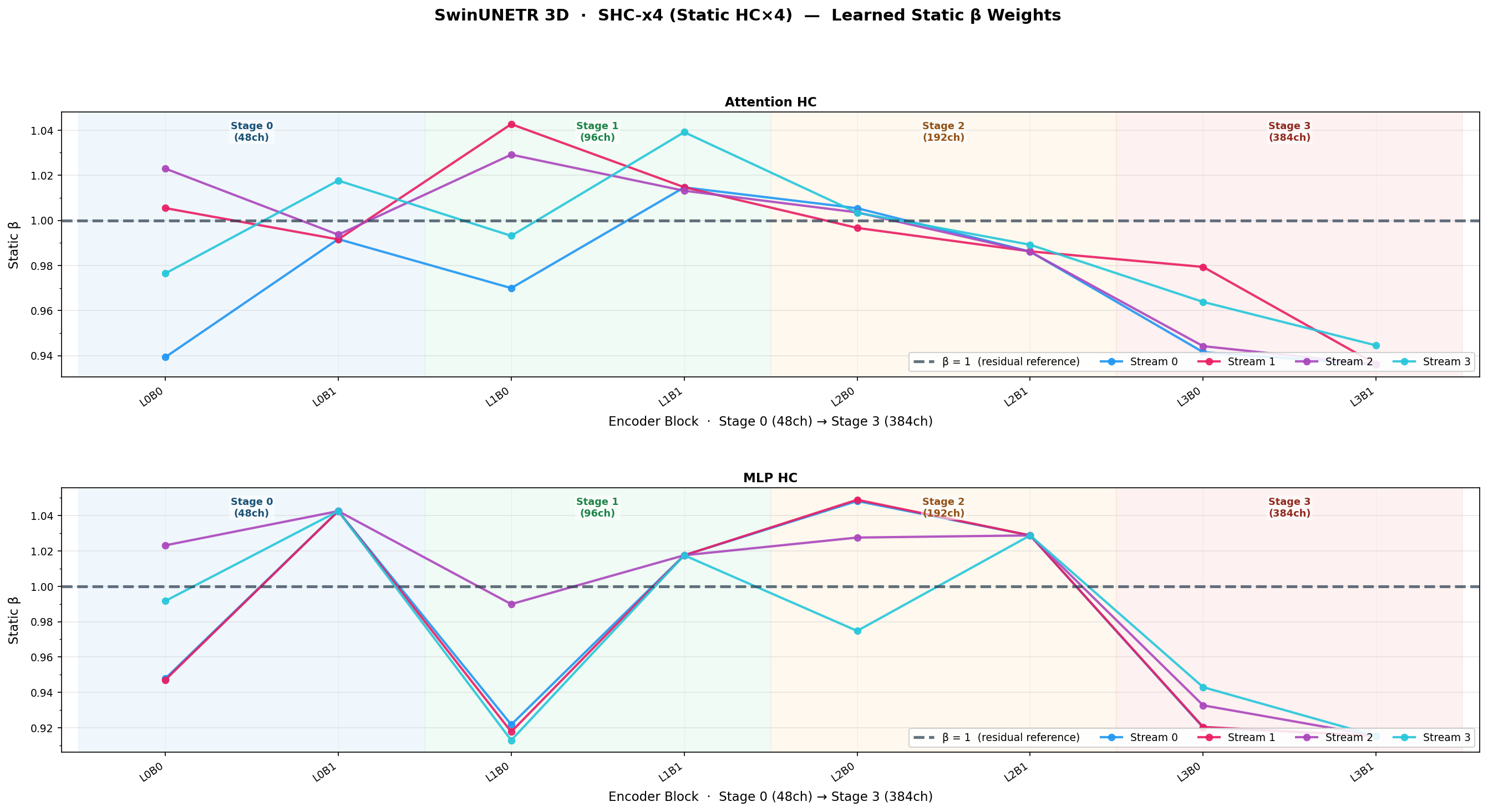}
\betaAtlas{VTUNet3D}{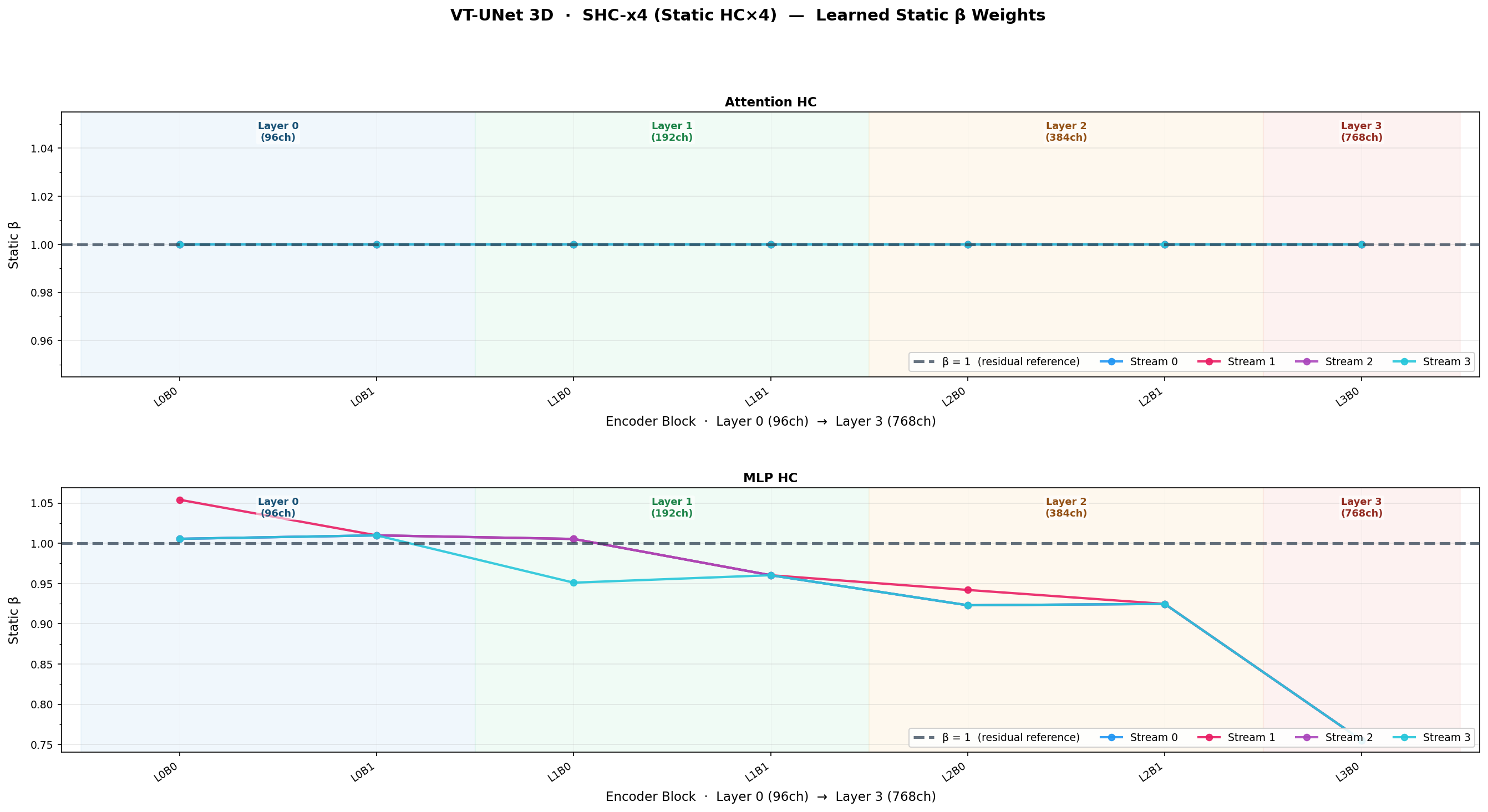}

\end{document}